\documentclass[11pt, a4paper, logo, onecolumn]{googledeepmind}

\pdftrailerid{redacted}
\usepackage{bbding}
\usepackage{wrapfig}

\usepackage{xspace}

\usepackage{colortbl}
\definecolor{minetable1colorx}{rgb}{0.75, 0.75, 0.75}

\definecolor{cvprblue}{rgb}{0.21,0.49,0.74}
\usepackage[pagebackref,breaklinks,allcolors=cvprblue, colorlinks=true]{hyperref} 
\usepackage{multirow}
\usepackage{multicol}
\usepackage{gensymb}
\usepackage{graphicx}
\usepackage{caption}

\definecolor{figtextcolor}{HTML}{333333}
\definecolor{perceptioncol}{HTML}{1A73E8}
\definecolor{modelingcol}{HTML}{925BCF}
\definecolor{manipulationcol}{HTML}{C440A7}
\definecolor{reasoningcol}{HTML}{DA2F77}

\usepackage{subcaption}
\usepackage{opensans}
\DeclareCaptionFont{subfigcapsize}{\fontsize{8pt}{8pt}\selectfont}  %
\usepackage{xurl}
\usepackage[capitalise]{cleveref}
\crefname{section}{Sec.}{Secs.}  %
\crefname{subsection}{Sec.}{Secs.}  %
\crefname{subsubsection}{Sec.}{Secs.}  %
\crefname{appendix}{App.}{Apps.}  %

\usepackage{graphicx}
\graphicspath{{figures/}}

\usepackage[breakable]{tcolorbox}

\newcounter{takeaway}
\usepackage{tikz}
\usetikzlibrary{calc}
\newlength{\tikzwidth}
\newlength{\boxpad}
\newlength{\boxgap}
\newlength{\boxtextheight}
\usepackage[numbers, sort&compress]{natbib}
\correspondingauthor{my-email@google.com}
\newcommand{\ModelName}{GenCeption\xspace}
\title{Video Generation Models are General-Purpose Vision Learners}

\reportnumber{} %

\author[1]{Letian Wang$^{1,2}$}
\author[1]{Chuhan Zhang$^{1}$}
\author[1]{Rishabh Kabra$^{1,3}$}
\author[1]{Jasper Uijlings$^{1}$}
\author[1]{Steven Waslander$^{2}$}
\author[1]{Andrew Zisserman$^{1,4}$}
\author[1]{Joao Carreira$^{1}$}
\author[1]{Kaiming He$^{1,5}$}
\author[1]{Misha Andriluka$^{1}$}
\author[1]{Eduard Gabriel Bazavan$^{1}$}
\author[1]{Andrei Zanfir$^{1}$}
\author[1]{Cristian Sminchisescu$^{1,6,*}$}

\affil[1]{$^{1}$Google DeepMind}
\affil[2]{$^{2}$University of Toronto}
\affil[3]{$^{3}$University College London}
\affil[4]{$^{4}$University of Oxford}
\affil[5]{$^{5}$MIT}
\affil[6]{$^{6}$Lund University}

\begin{abstract}

Driven by next-token prediction, NLP shifted from task-specific models into powerful generalist foundation models.
What, then, is the equivalent catalyst needed to achieve a general-purpose model in computer vision?
  In this paper, we contend that large-scale text-to-video generation serves as a strong pre-training paradigm for computer vision, providing the necessary spatiotemporal priors, vision-language alignment, and scalability required for general visual intelligence. We introduce \ModelName, which leverages a pre-trained video generative diffusion backbone to define a feed-forward perception model, capable of performing various vision tasks steered by text instructions. 
    Empirical results demonstrate that \ModelName achieves state-of-the-art performance across a diverse suite of tasks, including depth, surface normal, and camera pose estimation, expression-referring segmentation, and 3D keypoint prediction, often matching or surpassing specialized models 
    (e.g. DepthAnything3, SAM3, D4RT, VGGT-$\Omega$, Sapiens, David, Genmo, and Lotus-2).
    Furthermore, the video generative pretrained backbone outperforms alternative pretraining paradigms (e.g., V-JEPA, and Video MAE) under comparable settings.
    Importantly, \ModelName exhibits preliminary data and model scaling properties along with exceptional data efficiency where it achieves comparable performance with leading models like D4RT and VGGT-$\Omega$ with 7$\times$ to 500$\times$ less training data.
    Finally, \ModelName also exhibits intriguing emergent behaviors: a model trained exclusively  on synthetic human videos generalizes to real-world footage and out-of-distribution object categories (e.g., animals and robots). These findings suggest that video generation is not merely a synthesis tool, but a foundational path toward generalist vision intelligence for the physical world.
    \newline
    
    Project page: \url{https://genception.github.io}
\end{abstract}

\begin{document}

\maketitle
\makeatletter
\fancyhead[C]{\fontsize{7}{8}\selectfont\@title}
\makeatother

\section{Introduction}
\label{sec:intro}
Natural language processing (NLP) evolved from an era of specialized models—where separate models were developed for each language task (e.g. translation, summarization)—to a singular, unified foundation model paradigm. This paradigm, supported by large-scale next-token prediction pre-training followed by task-aligned post-training, has effectively collapsed thousands of disparate linguistic challenges into a single generalist intelligence, ultimately unlocking emerging behaviors such as chain-of-thought and in-context learning.

In stark contrast, computer vision is still lingering in its "specialized model" stage. While recent years have yielded powerful foundation models such as Segment Anything series~\cite{kirillov2023segment,ravi2024sam,carion2025sam} for localization or Depth Anything series~\cite{yang2024depth,yang2025depth,lin2025depth} for geometry, these remain fundamentally task-specific models that need customized architectures for each task. We have mastered specialized perception, but we have yet to realize a unified vision foundation model—a unified task-agnostic architecture that mirrors LLMs from general pre-training to versatile, emerging intelligence.
To this end, we posit that the quest for a generalist vision model is essentially a search for a universal pre-training objective—a visual analog to next-token prediction. Such a pre-training paradigm should satisfy three core imperatives:
\begin{enumerate}[label=\arabic*)]
\item \textit{Spatio-Temporal Evolution}: The world is a 4D continuum. The pre-training objective should force the model to internalize the 4D temporal causality and physics of a world in motion.
\item \textit{Vision-Language Alignment}: To inherit the instruction-following capabilities and common-sense knowledge of language models, visual features should be natively aligned with linguistic semantics during both pre-training and post-training stages.
\item \textit{Scaled up}: The paradigm should be scaled in both data and compute, ultimately enabling the emergence of vision intelligence.
\end{enumerate}

In this paper, we demonstrate that large-scale text-to-video generation serves as this pivotal pre-training paradigm, uniquely satisfying all three requirements, compared to prior vision pretraining methods (e.g. contrastive learning, masked autoencoder). First, the objective of generating high-fidelity video sequences forces a model to internalize the spatio-temporal world priors, including 3D geometry, object permanence, and physical interactions. Second, as these models are natively conditioned on text, they provide the inherent vision-language alignment. Finally, modern text-to-video generation models have been scaled using massive datasets and compute due to their relatively low annotation cost and high commercial value, providing a direct path toward emergent intelligence.

\begin{figure*}[tb]
  \centering
    \vspace{-1em}
   \includegraphics[trim=1cm 2.5cm 1cm 1cm, clip, width=1\textwidth]{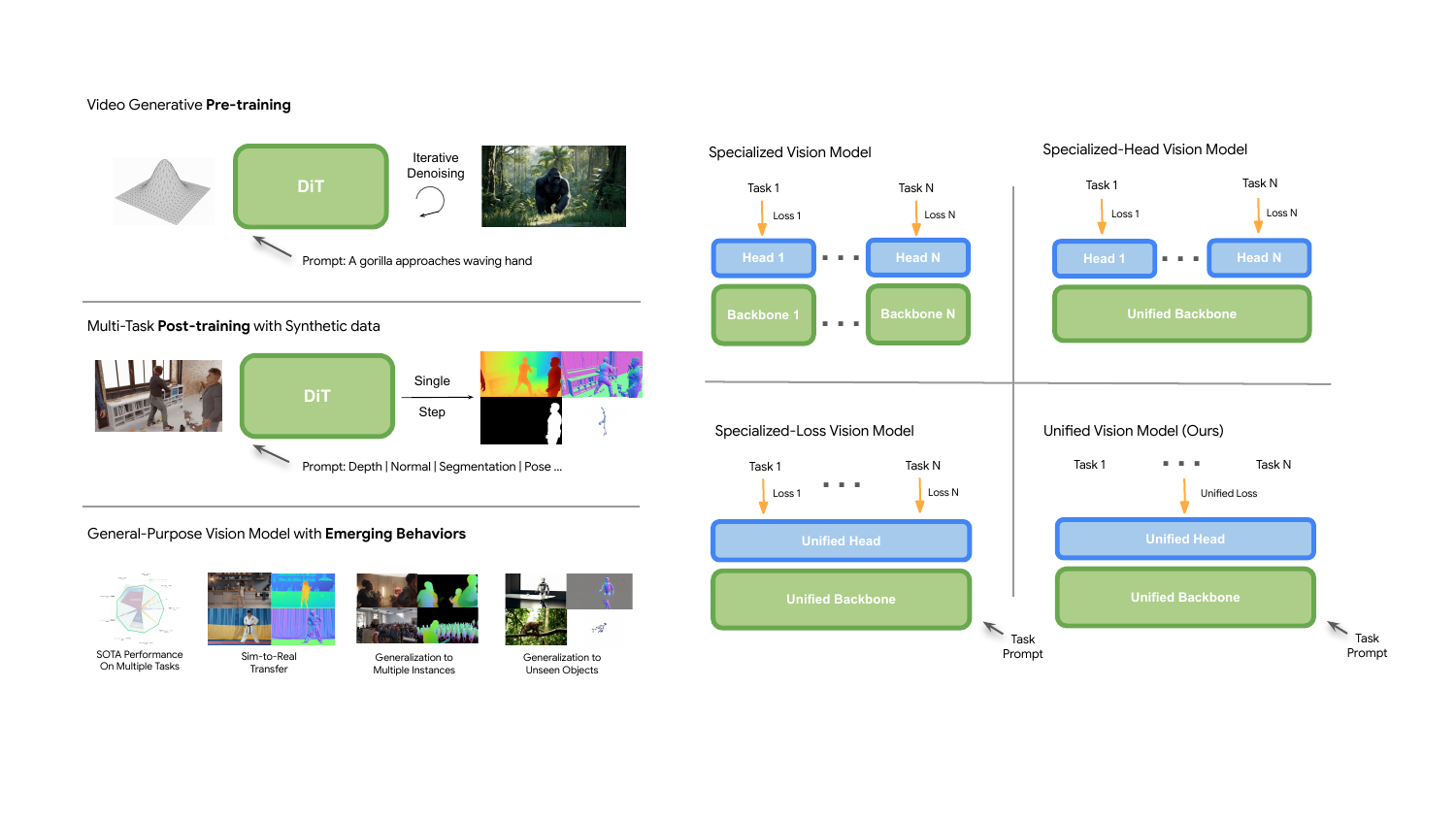}
  \caption{ 
  \textbf{Methdology} ({Left}): \ModelName treats a video generative diffusion model as a \textit{pre-training} base to capture rich spatio-temporal world priors and native vision-language alignment at scale.
  During multi-task \textit{post-training}, the model is adapted to feed-forward model fine-tuned on predominantly synthetic data to handle diverse perception tasks. 
  \ModelName shows strong performance with intriguing \textit{Emerging Behaviors}, enabling seamless sim-to-real transfer and generalization to out-of-distribution object categories.
  \textbf{Paradigm Shift} (Right): This highlights a paradigm shift from specialized task-specific computer vision models, to fully unified generalist vision models.
    }

  \label{fig:methodology}
\end{figure*}

\begin{figure*}[tb]
  \centering
   \includegraphics[trim=0cm 0cm 0cm 0cm, clip, width=1\textwidth]{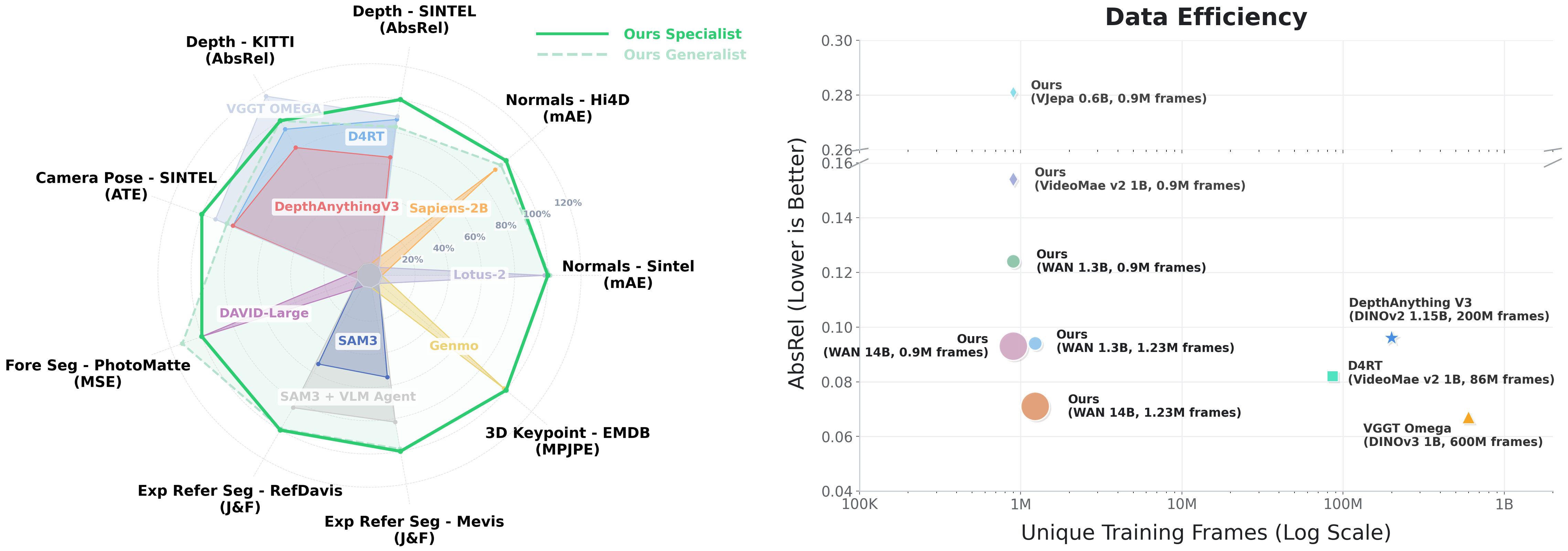}
   
  \caption{ 
  \textbf{SOTA Generalist Capability} (Left): \ModelName achieves universally competitive performance on a wide range of vision tasks, matching or outperforming state-of-the-art models dedicated to individual tasks (e.g. DepthAnything3~\cite{lin2025depth}, SAM3~\cite{carion2025sam}, D4RT~\cite{zhang2025efficiently}, VGGT-$\Omega$~\cite{wang2026vggt}, Sapiens~\cite{sapienseccv2024}, David~\cite{saleh2025david}, Genmo~\cite{genmo2025}, Lotus-2~\cite{he2025lotus}). 
  Our specialist denotes a model trained on each task individually, whereas the generalist represents a single model trained jointly across multiple tasks.
  \textbf{Data Efficiency in Finetuning} (Right): Validated on depth estimation, the video generative pretrained backbone 
  (i) outperforms the largest available variants alternative pretraining paradigms (e.g., V-JEPA, and VideoMAE V2) under the same finetuning data.
  (ii) exhibits preliminary scaling properties, where the performance improves with more data and large model size; 
  (iii) shows exceptional data efficiency, achieving comparable performance with leading models like D4RT~\cite{zhang2025efficiently} and VGGT-$\Omega$~\cite{wang2026vggt} with 7$\times$ to 500$\times$ less training data.
    }

  \label{fig:radar}
\end{figure*}

We introduce \ModelName, a generalist video perception model that instantiate this paradigm. As illustrated in Figure~\ref{fig:methodology}, we treat the pre-trained video diffusion backbone as our "base model", leveraging the rich priors learned during generative pre-training. We then apply the post-training phase, where we fine-tune the model across diverse visual tasks primarily with synthetic data. This enables \ModelName to execute heterogeneous perception tasks within a single task-agnostic architecture, ranging from pixel-level segmentation to more complex 3D pose estimation and tracking.
Specifically, by repurposing iterative diffusion into a feed-forward architecture, \ModelName dynamically infers the intended modality in a single forward pass, steered by the text instruction.
Notably, \ModelName\ promotes a paradigm shift from specialized vision models to unified generalist intelligence that adopt unified backbone, head, and loss function across different vision tasks.
Under this paradigm, task specifications are shifted from architectural modifications to data format designs, which enables scaling up with an ever-growing spectrum of visual tasks.

As shown in Figure~\ref{fig:radar}, our empirical results demonstrate that \ModelName achieves state-of-the-art performance across multiple perception tasks, performing on par with or even surpassing specialized models designed for single modalities, including DepthAnything V3~\cite{lin2025depth}, SegmentAnything V3 (SAM3)~\cite{carion2025sam}, D4RT~\cite{zhang2025efficiently}, VGGT-$\Omega$~\cite{wang2026vggt}, Sapiens~\cite{sapienseccv2024}, David~\cite{saleh2025david}, and Genmo~\cite{genmo2025}. 
Under comparable model size and same fine-tuning data, the video generative pretrained backbone is also found to outperforms the largest available variants of alternative pretraining paradigms (e.g., V-JEPA~\cite{bardes2024revisiting}, VideoMAE V2~\cite{tong2022videomae}).
Importantly, \ModelName exhibits preliminary scaling properties where the performance improves with more data and larger model size, along with exceptional data efficiency where it achieves comparable performance with leading models like D4RT and VGGT-$\Omega$ with 7$\times$ to 500$\times$ less training data.
Finally, this paradigm triggers emergent behaviors, including seamless sim-to-real transfer and out-of-distribution generalization to unseen object categories, as shown in Figure~\ref{fig:oodvis} and Figure~\ref{fig:multipeople}.
This evidence suggests that video generation is not merely a synthesis tool, but the foundational path toward generalist vision intelligence for the physical world.

\section{Related Work}
\subsection{Perception Foundation Models.}
Perception foundation models have recently attracted significant attention, with frameworks such as Segment Anything series~\cite{kirillov2023segment,ravi2024sam,carion2025sam} and Depth Anything series~\cite{yang2024depth1,yang2024depth,lin2025depth} leveraging massive datasets to achieve robust performance across diverse visual scenarios. Parallel efforts have sought to develop unified models capable of handling multiple tasks simultaneously. However, these works operate primarily in the image domain~\cite{saleh2025david,mizrahi20234m,bachmann20244m}, lacking an inherent understanding of temporal dynamics. Furthermore, multitasking in these models is often implemented at a rigid architectural level—relying on task-specific encoder, decoder, or losses~\cite{saleh2025david}—which lacks the flexibility required for truly general-purpose systems.
Even among models that attempt to process video and flexibly handle varied tasks~\cite{zhu2022uni,huang2025unityvideo}, they have not yet reached the level of efficacy seen in specialized, task-specific models when evaluated on a wide range of video tasks.
This stands in stark contrast to Large Language Models (LLMs), which have demonstrated that a single unified architecture can flexibly master a vast array of tasks with human-level or even superhuman proficiency. A truly unified video perception model that matches the general-purpose proficiency of LLMs has yet to be realized.

\subsection{Visual Representation Learning}
At the heart of this challenge lies the long-standing task of representation learning: learning rich and robust visual representations to enable a wide range of vision tasks. In the literature, unsupervised and self-supervised learning have become core paradigms for representation learning, as they scale efficiently with large amounts of unlabeled data. Among representative approaches, masked autoencoders~\cite{he2022masked,pathak2016context} (MAE), inspired by masked language modeling~\cite{devlin2019bert,brown2020language}, learn to reconstruct missing image regions and have established a foundational framework for visual pretraining, where the learned representations can be effectively finetuned to various downstream vision tasks. 
Later works like VideoMAE~\cite{tong2022videomae} and RVM~\cite{zoran2025recurrent} have extended this to the video domain.
In parallel, another prominent self-supervised paradigm leverages feature-level self-distillation, as emplified by the DINO serires~\cite{caron2021emerging,oquab2023dinov2,simeoni2025dinov3}, 
where a student network is trained to match the feature distributions of a dynamically updated teacher network across different global and local views of the same image.
However, these vision-only methods inherently lack explicit multimodal vision-language alignment, lacking the semantic grounding required to link visual patterns to high-level concepts.
Another prominent paradigm, contrastive learning, aims to align semantically similar samples while separating dissimilar ones. By performing contrastive learning across vision and language modalities, models such as CLIP~\cite{radford2021learning} and SigLip~\cite{zhai2023sigmoid} learn powerful multimodal representations that unlock a wide range of tasks such as open-vocabulary classification and segmentation. 
Despite the vibrancy of this field, multimodal representation learning at the video level remains a crucial problem, presenting numerous challenges.
Chief among these is the severe limitation in model scaling; while language models routinely scale to hundreds of billions of parameters, prominent video representation learners remain orders of magnitude smaller. This bottleneck stems from, among other reasons, the prohibitive computational cost of training on dense video data, alongside the added complexity of modeling intricate temporal dynamics.

\subsection{Re-purposing Diffusion Models.}
Recently, an emerging direction is to leverage the rich features learned in pre-trained diffusion models, a path which our work follows. 
Early work in this area focused mainly on single-image prediction.
\textit{Marigold} \cite{Ke_2024_CVPR} demonstrated how a pre-trained image diffusion
model like Stable Diffusion \cite{rombach2022high} can be repurposed through fine-tuning on
synthetic data to act as a 
monocular depth estimator.
\cite{martingarcia2024diffusione2eft} improves the efficiency of prior diffusion-based depth
estimators \cite{Ke_2024_CVPR,fu2024geowizard} by aligning the time step with the noise level,
showing that a simple end-to-end fine-tuning approach can create a single-step, deterministic model
that is both fast and highly accurate. \textit{GenPercept} \cite{xu2024diffusion} conducts a systematic study and
identifies key components such as feature injection, decoding mechanisms, and training objectives,
that are critical for successfully adapting pre-trained diffusion models to dense perception tasks. 
\textit{Diception} \cite{zhao2025diception} demonstrates strong vision generalists capable of addressing multiple tasks, steering perception tasks using text prompts.

An inherent limitation of these single-image models, however, is their inability to produce temporally consistent predictions when applied to video sequences. 
To address this, \textit{BufferAnytime}~\cite{kuang2024bufferanytimezeroshotvideo} proposes to adapt the existing image diffusion model to the video domain by incorporating a temporal layer for depth and normal estimation problems. 
Going further, various works have explored directly employing native video diffusion models to address the temporal consistency issues. \textit{DepthCrafter}~\cite{hu2024depthcraftergeneratingconsistentlong} and \textit{NormalCrafter}~\cite{bin2025normalcrafter} incorporate alignment objectives with large-scale Vision Foundation Models, such as CLIP~\cite{radford2021learning} or DINO~\cite{oquab2023dinov2}, to facilitate training. 
\cite{yang2025depthvideoscalablesynthetic} explores training with scalable synthetic data to overcome the scarcity of video ground truth data. 
\textit{Geo4D}~\cite{jiang2025geo4d} adapts video diffusion models for 4D scene reconstruction.
\textit{DiffusionRenderer} \cite{liang2025diffusionrendererneuralinverseforward} leverage video diffusion priors to tackle the dual problems of inverse rendering and forward rendering. 
\textit{ReferEverything}~\cite{bagchi2025refereverything} repurpose video diffusion models for expression-referring open-world segmentation.
\cite{acuaviva2025generation} provides related evidence that video diffusion models encode reusable visual priors, demonstrating few-shot LoRA adaptation to diverse tasks via input-output video transition.
However, these advanced methods primarily concentrate on a single, often dense, vision task.
We present a unified architecture for both dense and sparse video perception, leveraging the rich prior knowledge and instruction-following strengths of text-to-video models to steer diverse tasks via text prompts. 
Additionally, we replace slow iterative sampling with an efficient, direct feed-forward architecture, with superior inference efficiency and performance that matches or exceeds SOTA specialists. 
This work advances our previous human-centric model, THFM~\cite{wang2026thfm}, into a unified general-purpose model, with broader task and benchmark evaluations, and more exhaustive analysis of model behaviors.

\begin{figure*}[t!]
   \centering
   \includegraphics[width=1\textwidth]{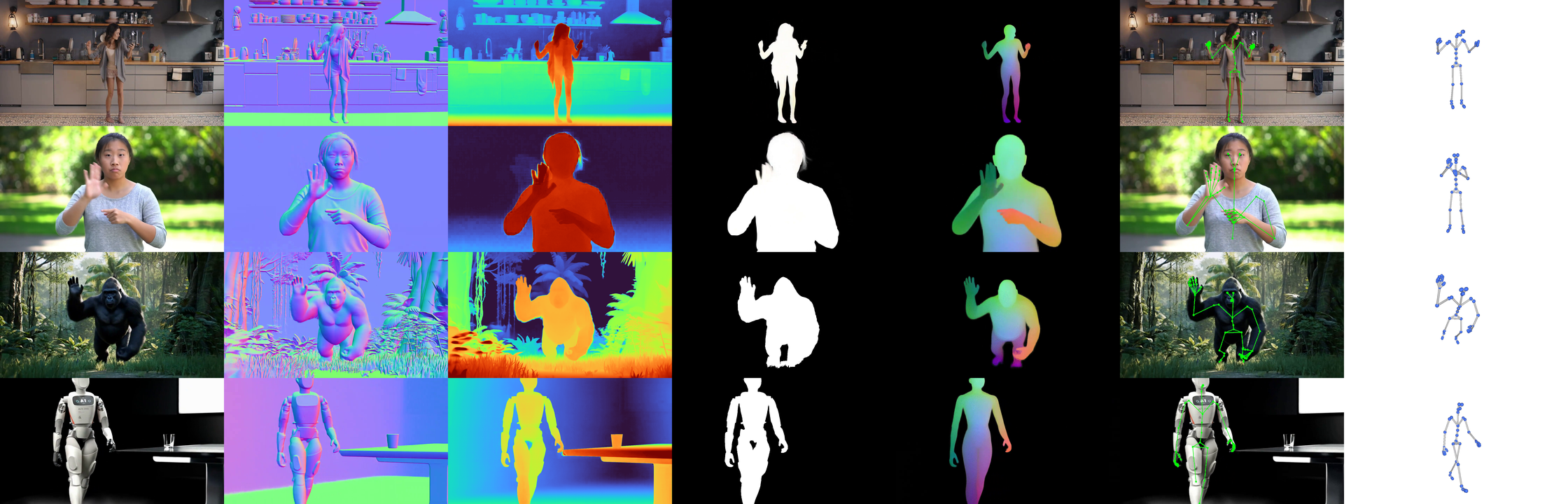}
   \captionof{figure}{\textbf{Capabilities Illustrations.} \ModelName is a versatile video perception model with SOTA performance on a multitude of output modalities including surface normal estimation, depth estimation, foreground segmentation, expression-referring open-world segmentation, dense human pose estimation, 2D/3D keypoints estimation, and camera pose estimation. Our approach has been trained on human-centric synthetic videos, yet it generalizes to real videos both for people as well as other categories such as animals and anthropomorphic characters. 
\label{fig:teaser}}
\vspace{0.5em}
\end{figure*}

Notably, our work shares close conceptual ties with two concurrent works, while differing in both architecture and evaluation. 
First, both our work and \cite{wiedemer2025video} share the hypothesis that large-scale video generative models learn powerful visual priors. If \cite{wiedemer2025video} demonstrates the existence of these priors, we provide the architecture to efficiently harness them. Specifically, while \cite{wiedemer2025video} investigates a training-free prompting approach via multi-step video generation with a focus primarily on qualitative validation, we introduces a dedicated post-training strategy and conduct rigorous quantitative evaluation across standardized benchmarks.
Second, while \textit{Vision Banana}~\cite{gabeur2026image} operates within the 2D image space, \ModelName extends the paradigm to the native video domain to naturally capture temporal consistency. Additionally, instead of focusing on multi-step generation, our method adopts a efficient feed-forward architecture with improved inference efficiency and strong performance across a broader task spectrum. See Figure \ref{fig:teaser}, Figure~\ref{fig:combined_demo}, Figure~\ref{fig:segmentation}, and Figure~\ref{fig:skiing} for more demonstrations of capabilities.

\begin{figure*}[tb]
  \centering
   \includegraphics[trim=0.5cm 5cm 1cm 1cm, clip, width=1\textwidth]{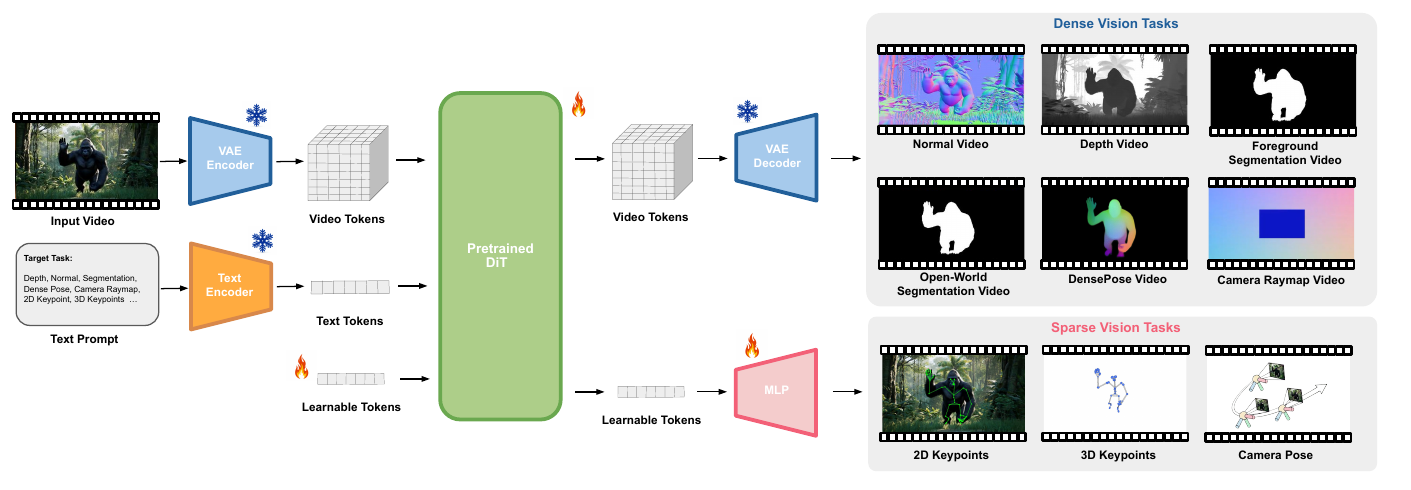}
  \caption{ 
  Architecture overview of \ModelName, a simple yet powerful architecture adapted from text-to-video diffusion models. 
  Given an input video and a text prompt specifying the desired output, our unified model, trained majorly on synthetic data, is capable of performing a wide range of dense and sparse perception tasks, with a single forward-pass of the model. 
   The dense vision tasks are unified in the RGB ambient space where supervision can be applied in latent space efficiently, and the sparse vision tasks are realized by adding learnable tokens as additional inputs to the diffusion transformer (DiT). 
    }

  \label{fig:overview}
\end{figure*}

\section{\ModelName}
\subsection{Methodology}
The core premise of \ModelName is that large-scale video generation models are not merely synthesis engines, but powerful, universal visual representation learners. Inspired by the trajectory of Large Language Models (LLMs)—where generative pre-training (next-token prediction) inherently builds rich features that can be harnessed for diverse downstream tasks—we reformulate video perception as a post-training endeavor built atop a pretrained video generative model. Our methodology is driven by three foundational principles:
\begin{enumerate}[label=\arabic*)]
    \item \textit{Multimodal Generative Pre-training as Representation Learning}: We utilize a pre-trained text-to-video generative diffusion model as our universal backbone. The objective of generating high-fidelity video conditional on text inherently forces the model to internalize robust spatiotemporal priors, 3D geometry, and physical laws, while simultaneously fostering a deep, aligned understanding of visual-language concepts.
    To fully preserve these foundational representations, one core design principle is to maximize compliance with the original pre-training regime with minimal modifications.
    \item \textit{Task-Agnostic Post-Training in Unified Architecture}: To transition from a specialized generator to a generalist perceiver, we treat diverse vision tasks as unified sequence-to-sequence mapping problems, rather than engineering task-specific architectures or relying on specialized encoders, decoders, and loss functions. By formatting various target modalities into a shared representational space (e.g., standard RGB space for dense tasks), we allow a single, unified architecture to master multiple tasks guided seamlessly by text instructions.
    Under this paradigm, integrating new visual capabilities no longer necessitates altering the architecture or training recipes; instead, task specifications are directly reflected in the data representation—a data-driven approach that enables scaling up efficiently across a continually expanding range of visual tasks.
    \item \textit{Feed-Forward Reformulation}: While diffusion models typically rely on slow, iterative sampling processes for synthesizing diverse outputs, perception tasks demand highly accurate and efficient predictions. Therefore, a critical design decision in \ModelName is the transformation of the multi-step generative backbone into a single-step, feed-forward perception model.
    
\end{enumerate}
Guided by these principles, we instantiate {\ModelName}, a model capable of various video perception tasks. As illustrated in Figure~\ref{fig:overview}, we repurpose text-to-video generation models to enable both dense vision tasks (normal estimation, depth estimation, foreground segmentation, expression referring open-world segmentation, dense human semantic estimation, raymap estimation) and sparse vision tasks (2D and 3D keypoint prediction) within a unified architecture. 
Given the input RGB video and the text prompt specifying the target modality, \ModelName is able to produce target video in a single forward pass (Sec~\ref{sec: architecture}), with a unified representation for various perception tasks (Sec~\ref{sec: representation});
A scalable data synthesis strategy is proposed to enable efficient low-cost collection of diverse high-quality data (Sec~\ref{sec: data synthesis}).
We then discuss our training recipe in Sec~\ref{sec: training recipe}.

\subsection{From Diffusion Pre-training to Perception Finetuning}
\label{sec: architecture}

{\textbf{{Preliminary}.}}
Our model is based on text-to-video diffusion model, with its core components being a VAE encoder-decoder pair, a text encoder, and a transformer-based latent diffusion model (DiT). 
The original DiT model takes a latent token initialized with Gaussian noise and text prompt as inputs and conducts a number of denoising steps, gradually transforming it into latent tokens, which can be decoded to the generated video. The index of the denoising step is provided to DiT as additional input to modulate the predictions with respect to amount of input noise. Let us denote the video token inputs to DiT at step $t$ as $x_t$, with $x_0$ being the noise-free latents of the original video. Depending on the particular formulation, DiT module is trained to predict the noise component $\epsilon$ as in \cite{song2020denoising,ho2020denoisingdiffusionprobabilisticmodels}, the velocity $v = \epsilon - x_0$ between the noise and original input as in the "Rectified Flow" variant \cite{liu2022flow, lipman2022flow, esser2024scaling}, or the target $x_0$ directly~\cite{li2026back}.

\noindent\textbf{{Feedforward Perception Formulation.}} 
In this work, we repurpose the multi-step diffusion transformer into a feed-forward prediction model. To that end, instead of taking noisy video latents as inputs, we directly feed the clean latent of the input video as the inputs to DiT, and conduct only a single forward-pass of the model. 
To signify that the input is noise-free, the conditioning timestep is fixed to $t=0$, corresponding to the termination of the generative Rectified Flow process.
Since the DiT is trained under the Rectified Flow objective to predict the velocity $v = \epsilon - x_0$, we thus negate the raw output of the DiT prior to passing it to the loss functions or decoder. The negated output $-v = x_0 - \epsilon$ aligns with the latent of the target video more closely, which, empirically accelerates model convergence and improves performance. 

\noindent An intuitive way to view this formulation is that we recast the DiT as a powerful feature extractor. Crucially, by leveraging Rectified Flow rather than traditional noise prediction, conditioning on $t=0$, and negating the velocity output, we strive to extract the most informative and richest features from the pretrained backbone.
Furthermore, we harvest features directly from the final layer rather than intermediate ones; this ensures they natively align with the subsequent decoder, without any architectural modifications to the base model.

\subsection{Unified Task Representation}
\label{sec: representation}

\begin{figure}[htbp]
  \centering
  \begin{minipage}[c]{0.5\textwidth}
    \centering
    \includegraphics[trim=4.5cm 2.5cm 9.5cm 5cm, clip, width=\textwidth]{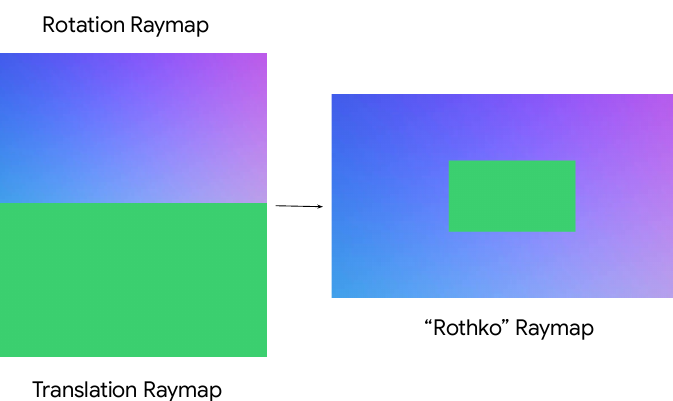}
  \end{minipage}
  \hfill
  \begin{minipage}[c]{0.45\textwidth}
    \caption{The "Rothko" Raymap as an example of adapting high-dimensional modal data into standard 3 RGB channels. This representation effectively compresses the camera's multi-channel ray data by assembling rotation and translation components into a single three-channel map.}
    \label{fig:raymap}
  \end{minipage}
\end{figure}

\noindent\textbf{{Unify Dense Tasks in RGB Space.}} 
Unlike prior practices that often adopt separate backbones, decoders, and loss functions for different tasks, our framework promotes a unified approach, where all tasks share the same architecture and weights for both backbone and decoder, and the target task is simply modulated via the text prompt.
To achieve this, the outputs for all dense perception tasks are represented in the 3-channel RGB space within the range $[0, 1]$. Specifically, 
the three RGB channels are replicated for single-dimensional outputs like depth and segmentation, whereas they represent distinct dimensions for three-dimensional tasks including surface normals and DensePose estimation. 
Importantly, this representation is also applicable to higher-dimensional vision modalities, many of which can be elegantly adapted to fit within this 3-channel constraint. 
For instance, to estimate camera poses which are traditionally represented as matrices, we employ a pixel-space raymap~\cite{zhang2024cameras,zhao2025diffusionsfm,jiang2025geo4d} that natively aligns with the video generative format (see Figure \ref{fig:raymap}). To fit its 6-channel format within our 3-channel constraint, we spatially partition the frame—allocating ray origins to the central region and ray directions to the periphery. This simple yet effective layout preserves the integrity of our single-decoder framework while fully leveraging pre-trained visual priors.
More importantly, this design philosophy scales naturally across the spectrum of vision tasks. Analogous to reformatting non-text data into text strings for LLMs, we advocate for projecting fundamentally visual tasks directly into the continuous pixel space—the native domain where pre-trained visual priors are best leveraged.

\noindent\textbf{{Enable Sparse Tasks with Learnable Tokens.}}
While RGB is versatile, downstream tasks like robotic control often require directly manipulatable, structured outputs such as explicit 2D/3D coordinates. To enable these sparse predictions, we introduce a lightweight token-based extension. We append $T$ learnable tokens—one for each video frame—to the video latents. After passing through the model, an MLP decodes each token to predict a $K$-dimensional target per frame. To maintain compliance with the base DiT model, we apply its native 3D RoPE to the additional tokens. Specifically, because the spatial positions of these additional tokens are unknown, we make them learnable. 
For temporal positioning, since the video frame count $T$ is compressed to a latent length $T'$, we down-scale the frame indices of the $T$ tokens using position interpolation~\cite{chen2306extending} to stay within the temporal bounds seen during pre-training.
Empirically, this additional-query approach outperforms methods that rely on adding extra attention layers.

\begin{figure*}[!tb]
\centering
  \includegraphics[width=1\textwidth]{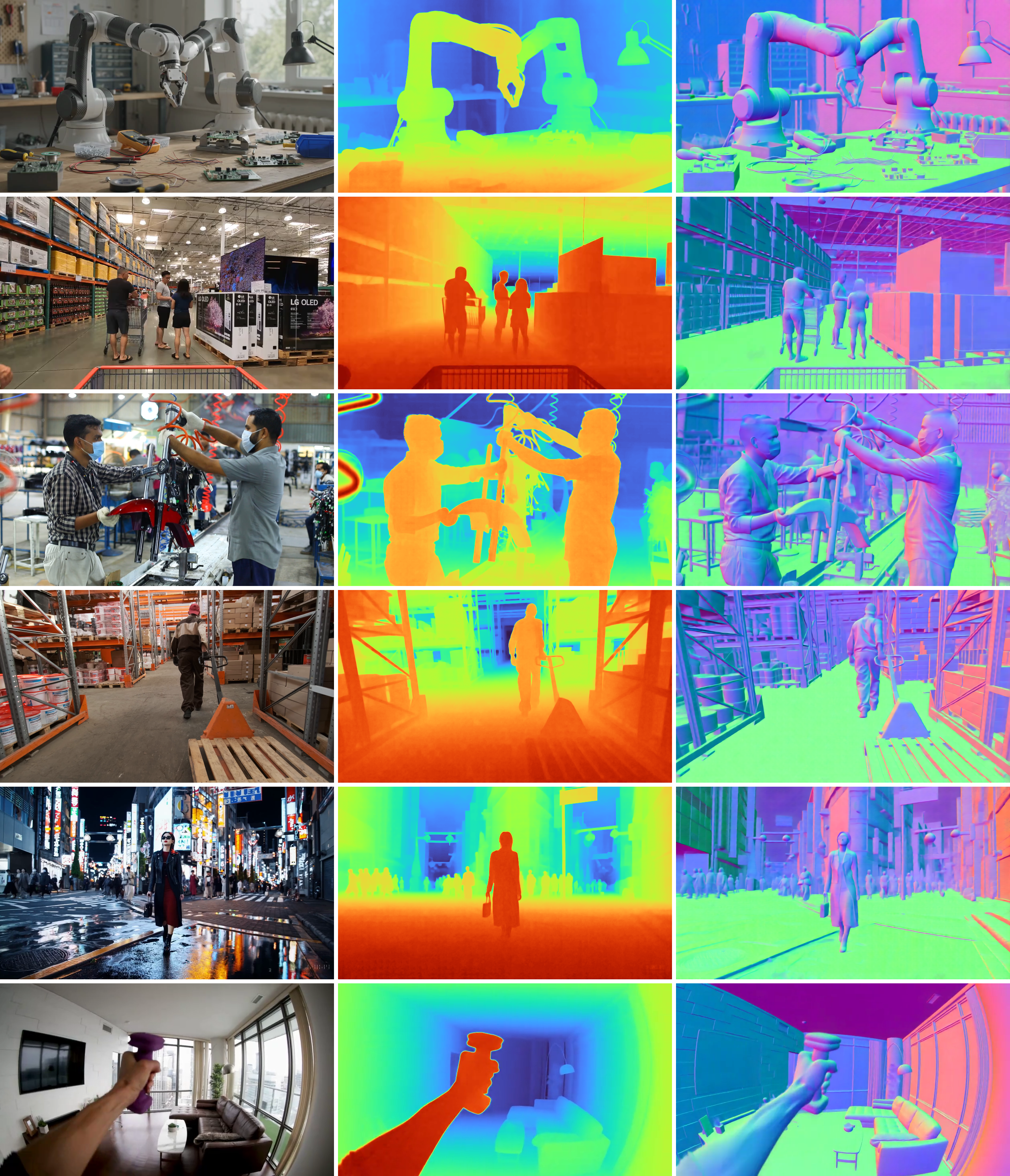}
  \caption{Demonstration of \ModelName’s depth and surface normal estimation capabilities. 
    }
  \label{fig:combined_demo}
    
\end{figure*}

\subsection{Scalable Synthetic Data Generation}
\label{sec: data synthesis}
Training of our model requires high-quality and diverse video data with ground-truth for depth, normal, segmentation mask, dense pose, 2D keypoints, 3D keypoints, and camera poses. Since most real-world datasets may contain only a subset of the modalities in a limited scale, we resort to synthetic data as a scalable approach to address the data scarcity problem. 

Specifically, we design a synthetic data generation workflow to produce high-quality human-centric video data, covering diverse human entities and motions. We use 800 RenderPeople assets~\cite{Renderpeople.com}, and animate them with 200 motions from the CMU motion capture dataset~\cite{CMUmocap}. 
We use various 3D full scenes or HDRI backdrops similar to \cite{bazavan2022hspacesyntheticparametrichumans} for enhanced background variability, and vary focal lengths, camera positioning and trajectory for enriched camera conditions.
In total, we generate a synthetic dataset of 7,500 videos with various identities, motions and backgrounds. 
To support dense vision tasks, video normals, depth and segmentation masks are generated with separate render passes via Blender \cite{BlenderOnlineCommunity2025}. 
We recover the human joint positions from the rigged RenderPeople \cite{Renderpeople.com} assets and use them as ground truth for our 2D and 3D keypoint regression task. Each of the generated videos is trimmed to the target number of frames in the video model. 
We apply a offline data preprocessing stage, where we precompute and cache both the input RGB video and target-modality video into video latents, and the text conditioning prompt into text embeddings.

\subsection{Training Recipe}
\label{sec: training recipe}

With the model architecture and data pipeline established, we now describe the training objectives employed in our framework.
Conventionally, joint multi-task learning presents an inherent optimization challenge: distinct vision tasks demand specialized loss functions tailored to their unique modalities. 
In practice, this heterogeneity introduces substantial engineering overhead in manually balancing loss weights across different objectives, presenting a major bottleneck when scaling up to a larger variety of vision tasks, data, and larger model sizes. 
Instead, we adopt a fully unified loss design: 
our model is trained solely using a standard $L_2$ loss, mirroring how LLMs unify diverse language tasks under a single objective. This single loss function is applied directly in the latent space for dense tasks and in the output space for sparse tasks.

To achieve this unified optimization without compromising performance, we shift the responsibility of task-specific customization from the loss formulation to the data representation designs. 
Taking monocular depth estimation as a primary example—where scale ambiguity typically necessitates customized, scale-invariant losses—we resolve this constraint directly at the data level. Specifically, we normalize the depth map of each video using the median depth of the scene, thereby inherently eliminating scale ambiguity. 
To further align the normalized depth $d$ into the standard $[0, 1]$ RGB range, similar to~\cite{gabeur2026image}, we apply a nonlinear mapping function: $d' = \text{clip}(\alpha \log(d + 1), 0, 1)$,
where $\alpha$ can dynamically adjusts the model's focus between near-field details and far-field structures. 
This data-level customization effectively eliminates the need for any task-specific objective functions, where balancing between different tasks only needs to be managed via the data mixture ratio, analogous to the prevailing practice in LLMs.
We believe this data-centric harmonization principle can scale naturally across a wider spectrum of vision tasks, ensuring a streamlined and robust multi-task training paradigm.

\begin{figure*}[tb]
\centering
  \includegraphics[width=1\textwidth]{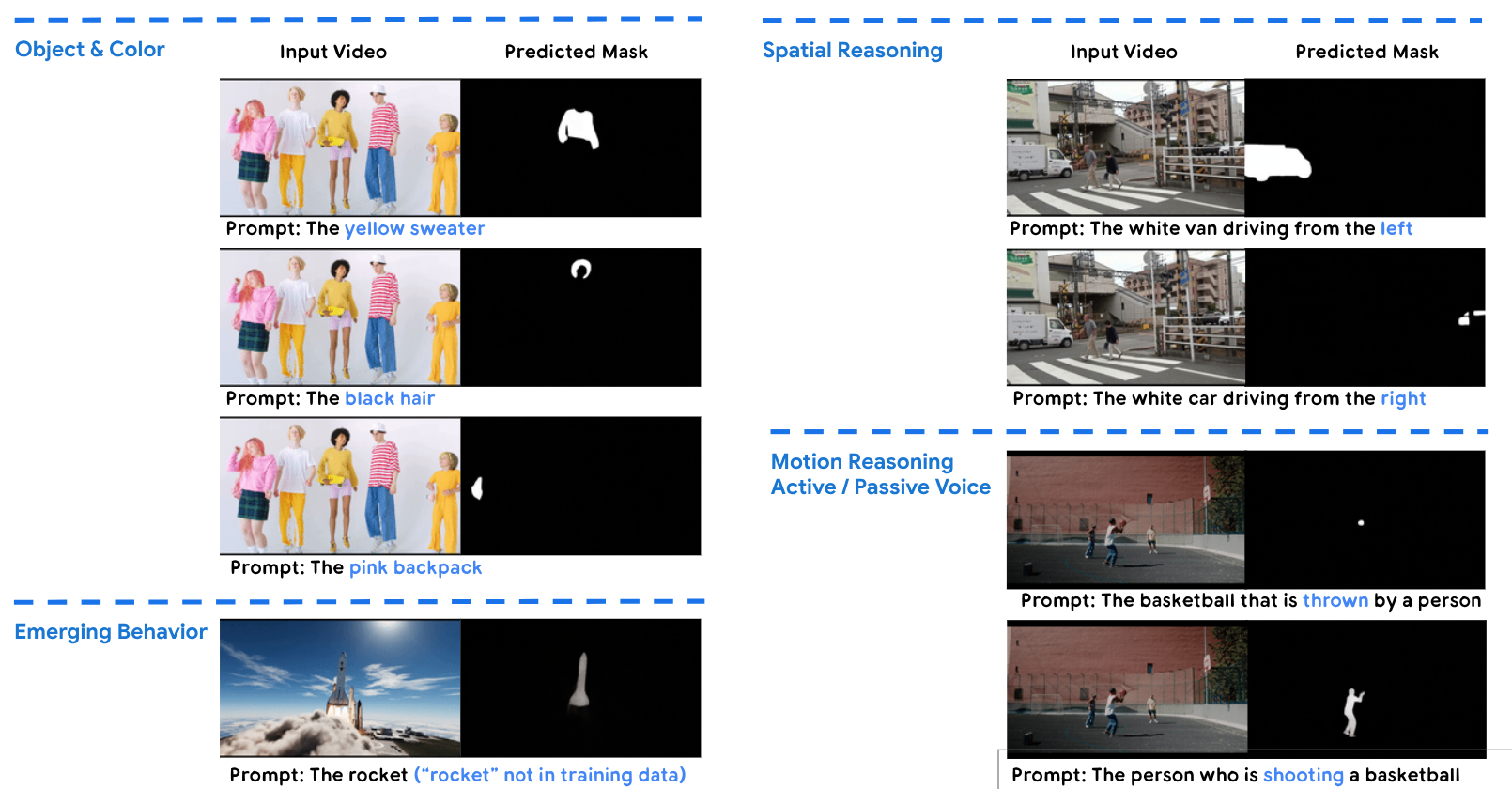}
  \caption{Qualitative results of our approach on referring expression segmentation. Our model accurately recognizes objects, colors, spatial relationships, and motion. Furthermore, it demonstrates strong generalization by identifying unseen objects, effectively leveraging the knowledge acquired during text-to-video pretraining.
    }
  \label{fig:segmentation}
\end{figure*}

\section{Experiments}
We comprehensively evaluate our model on various challenging real-world datasets across vision tasks: depth estimation, normal estimation, segmentation, 3D keypoint prediction, and camera pose estimation.  

\begin{table}[tb!]
    \centering
    \begin{tabular}{ccc}
        \includegraphics[width=0.3\textwidth]{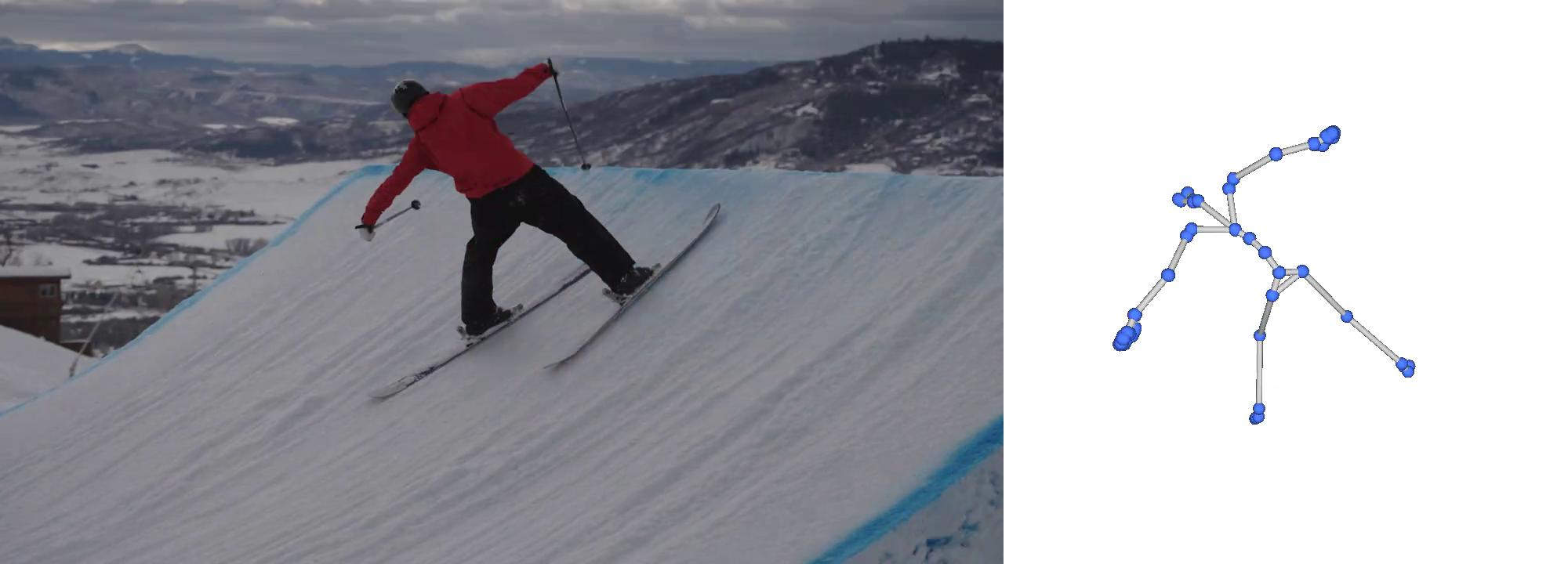} & 
        \includegraphics[width=0.3\textwidth]{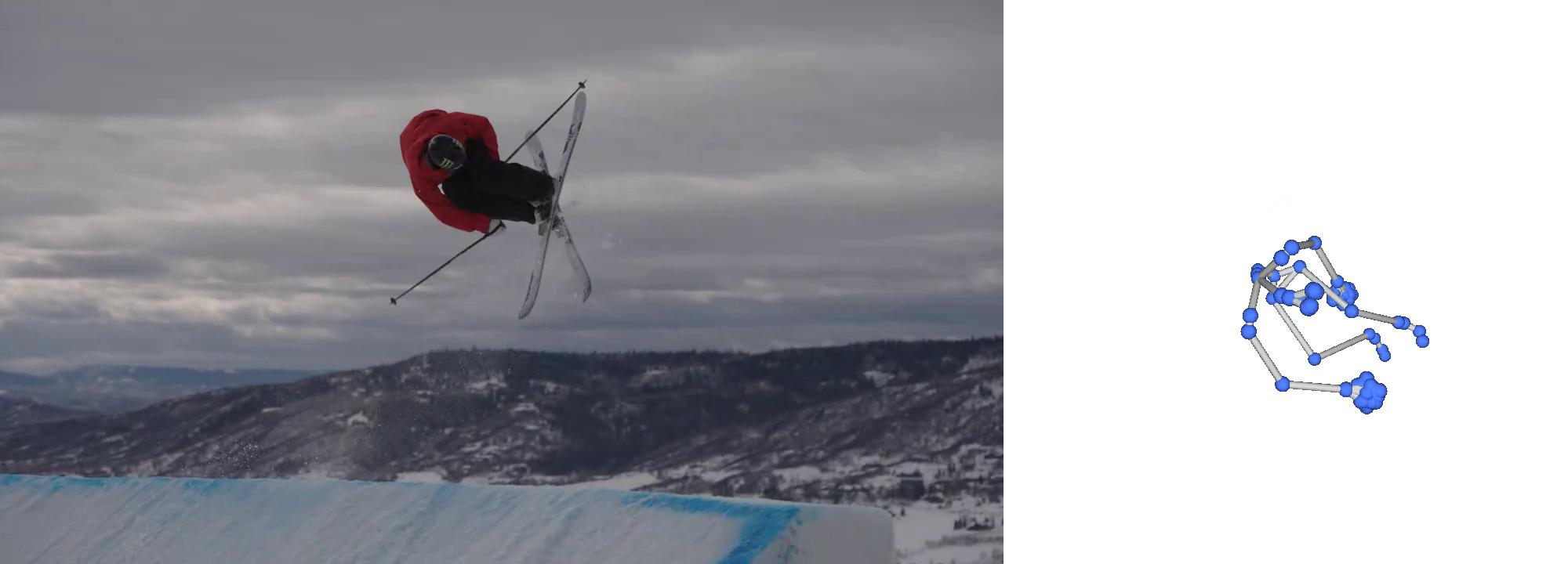} & 
        \includegraphics[width=0.3\textwidth]{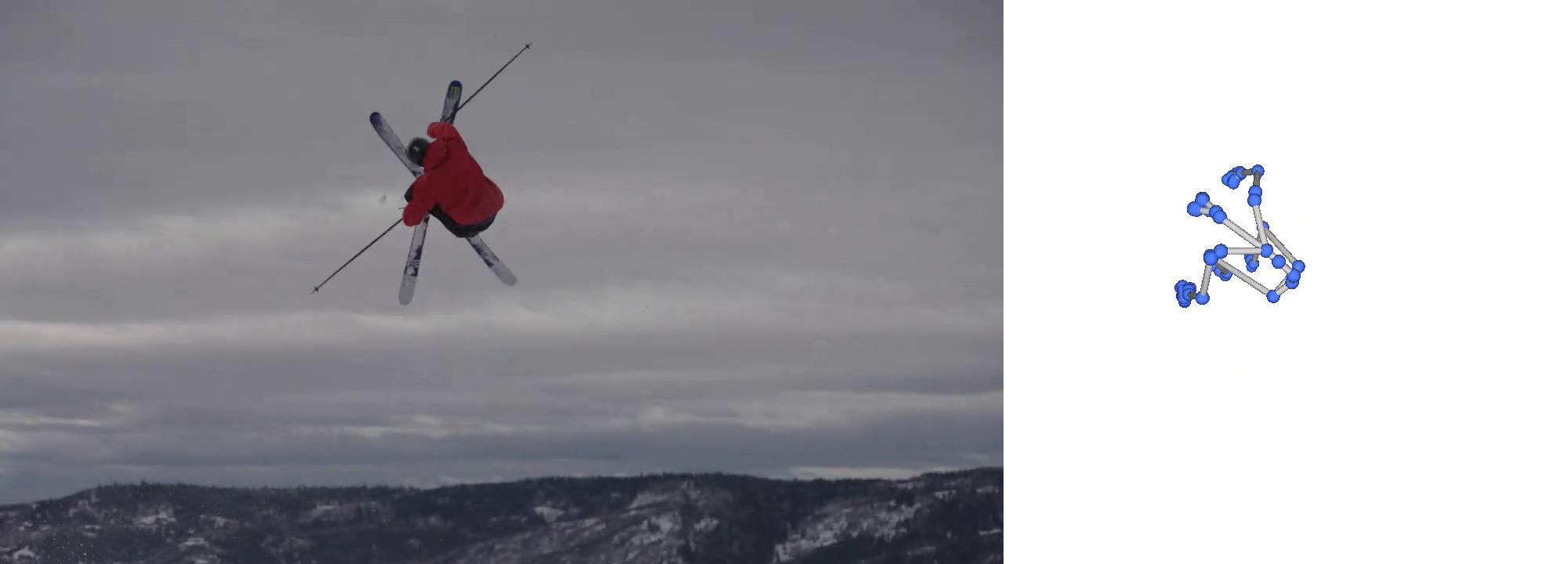} \\
        
        \includegraphics[width=0.3\textwidth]{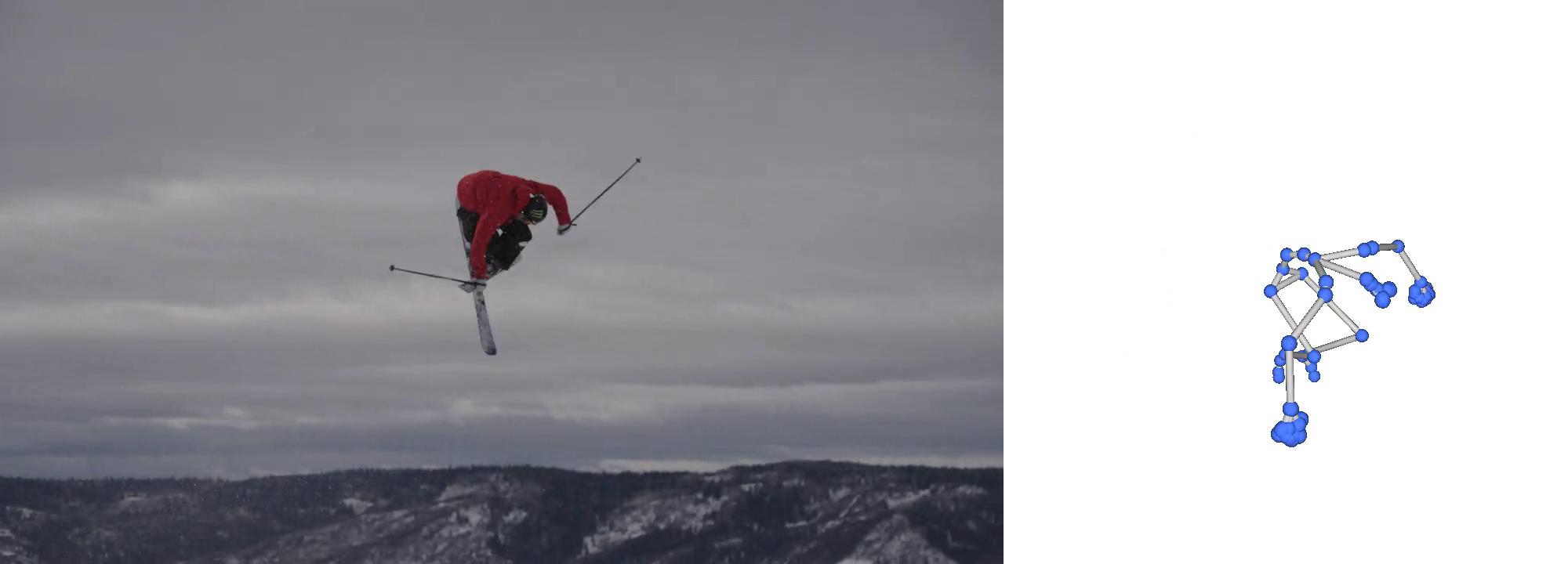} & 
        \includegraphics[width=0.3\textwidth]{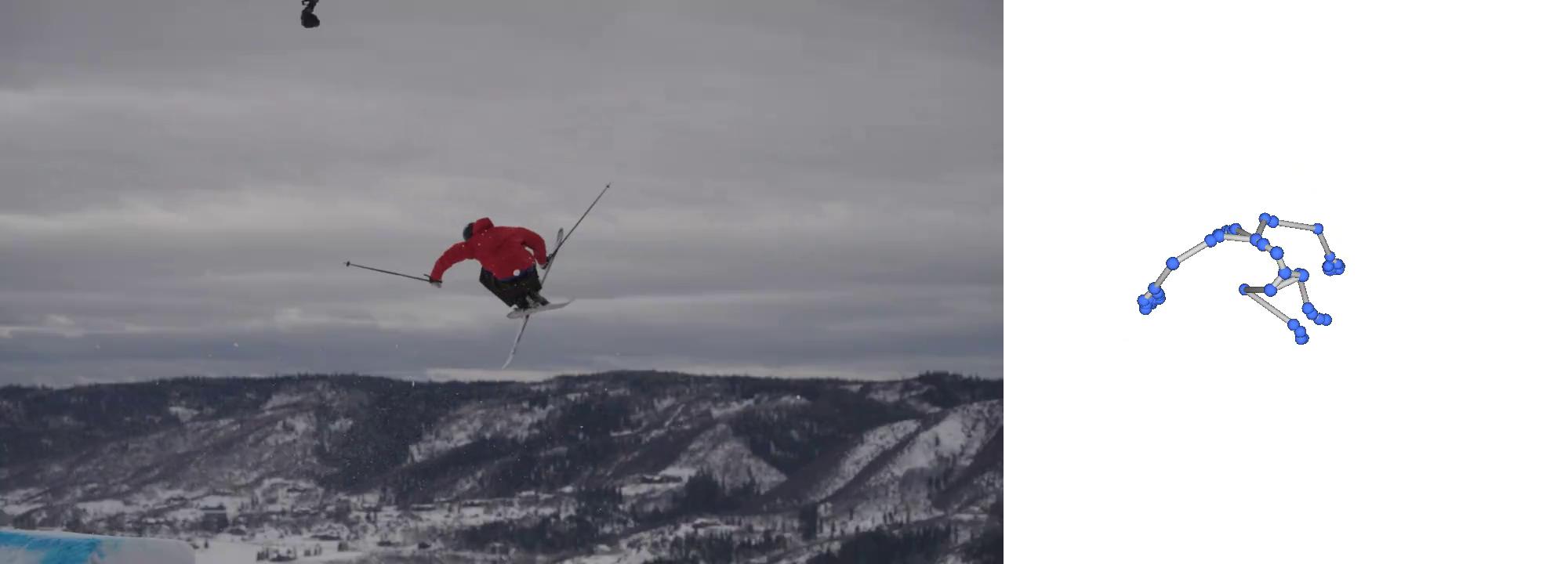} & 
        \includegraphics[width=0.3\textwidth]{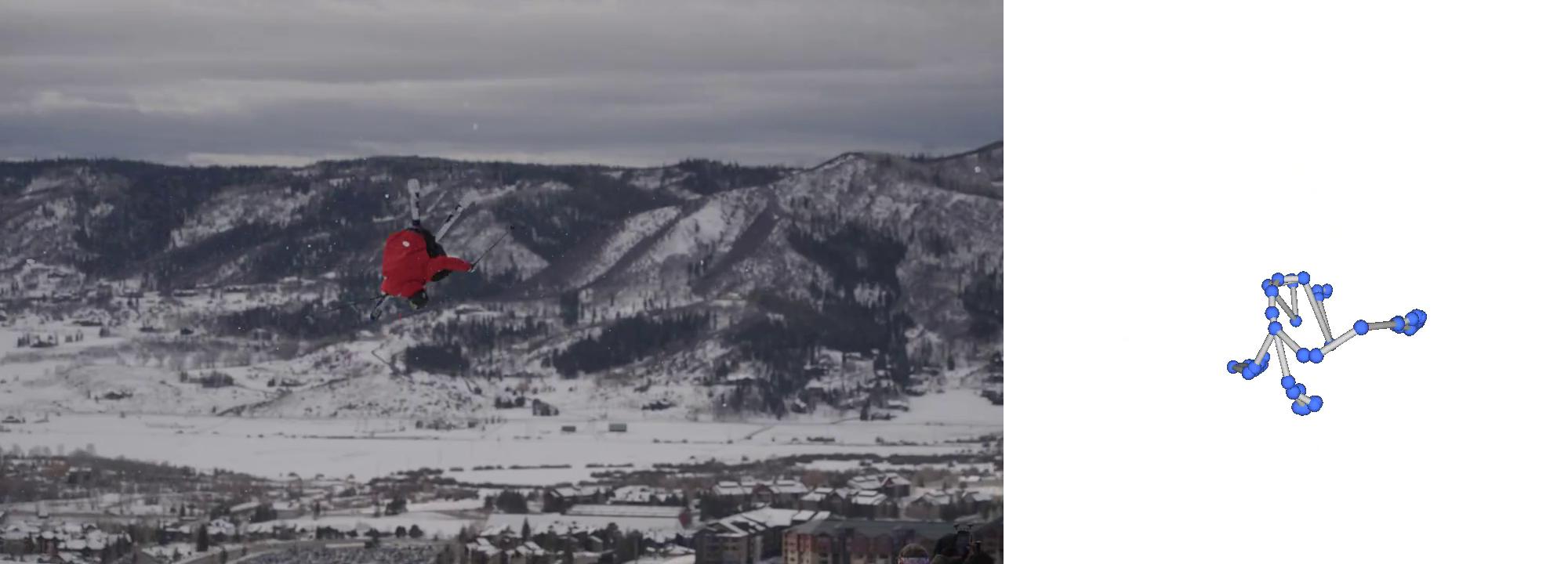} \\  
        
        \includegraphics[width=0.3\textwidth]{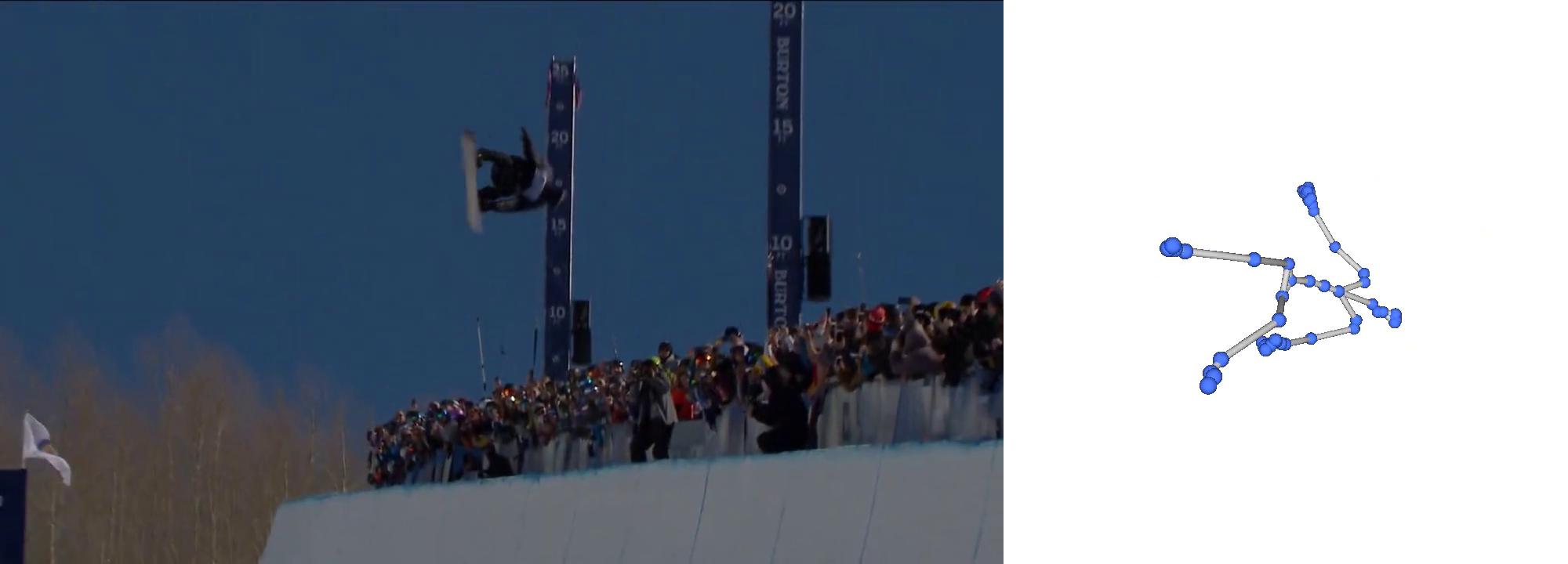} & 
        \includegraphics[width=0.3\textwidth]{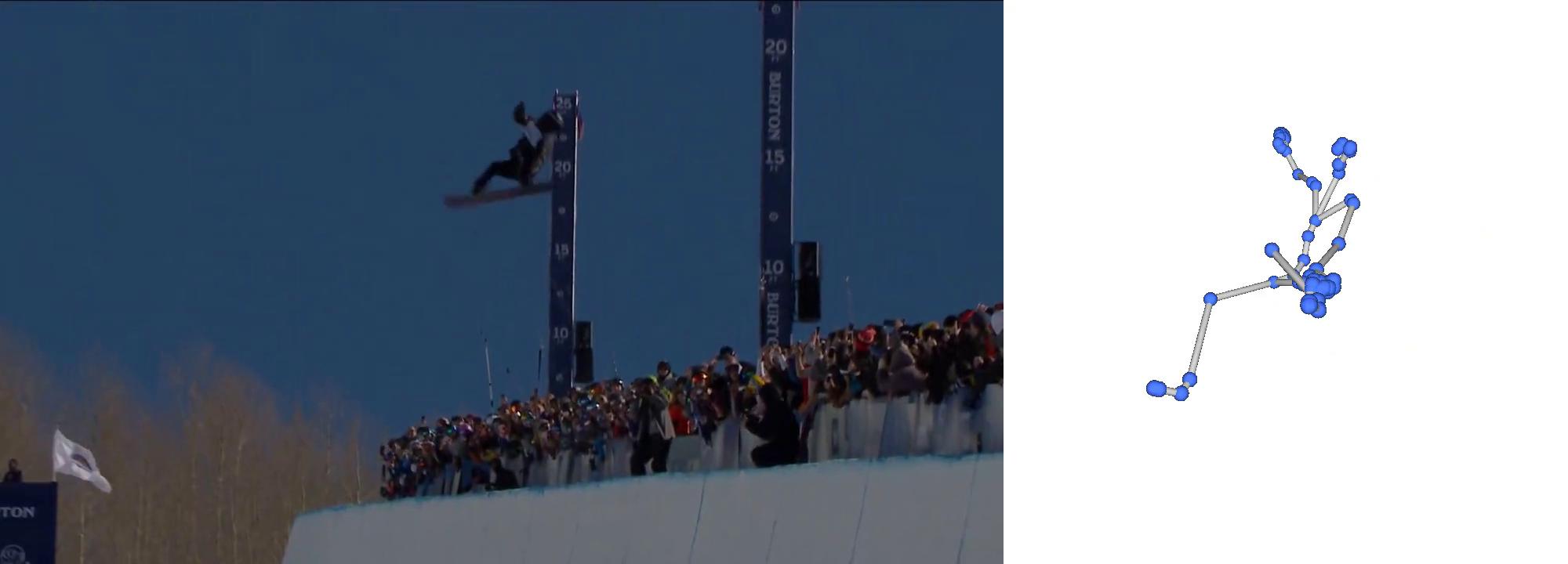} & 
        \includegraphics[width=0.3\textwidth]{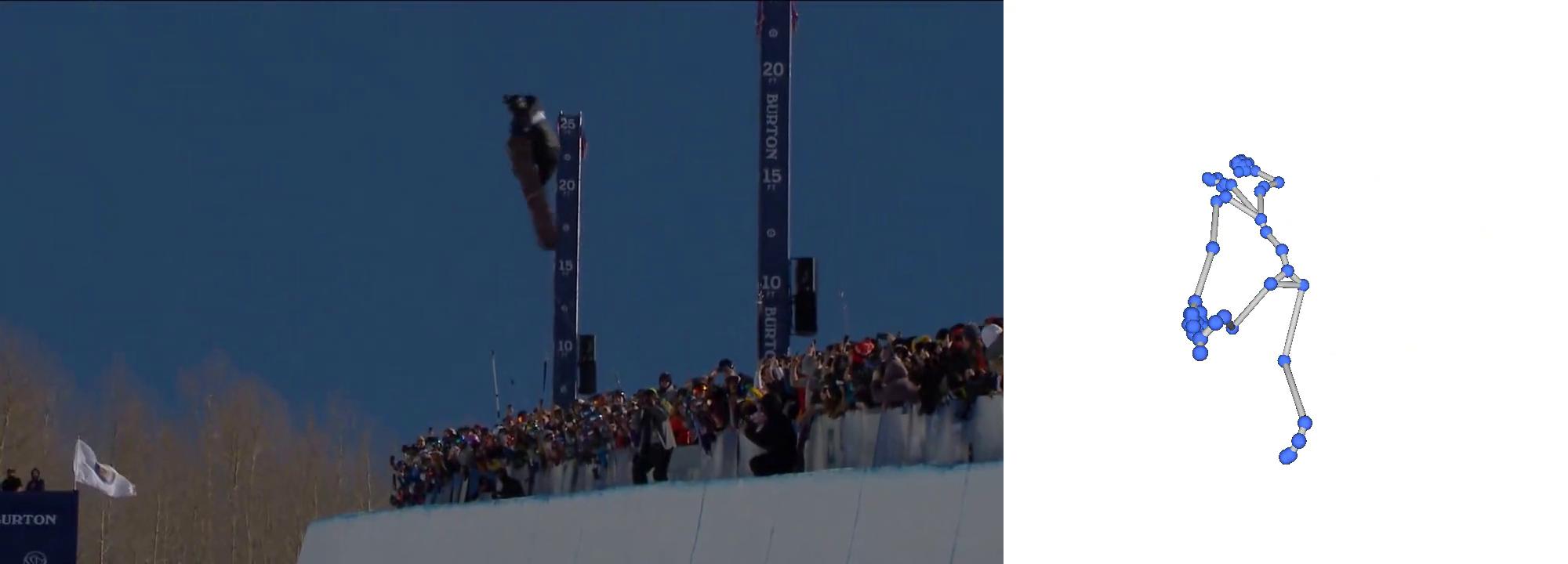} \\

        \includegraphics[width=0.3\textwidth]{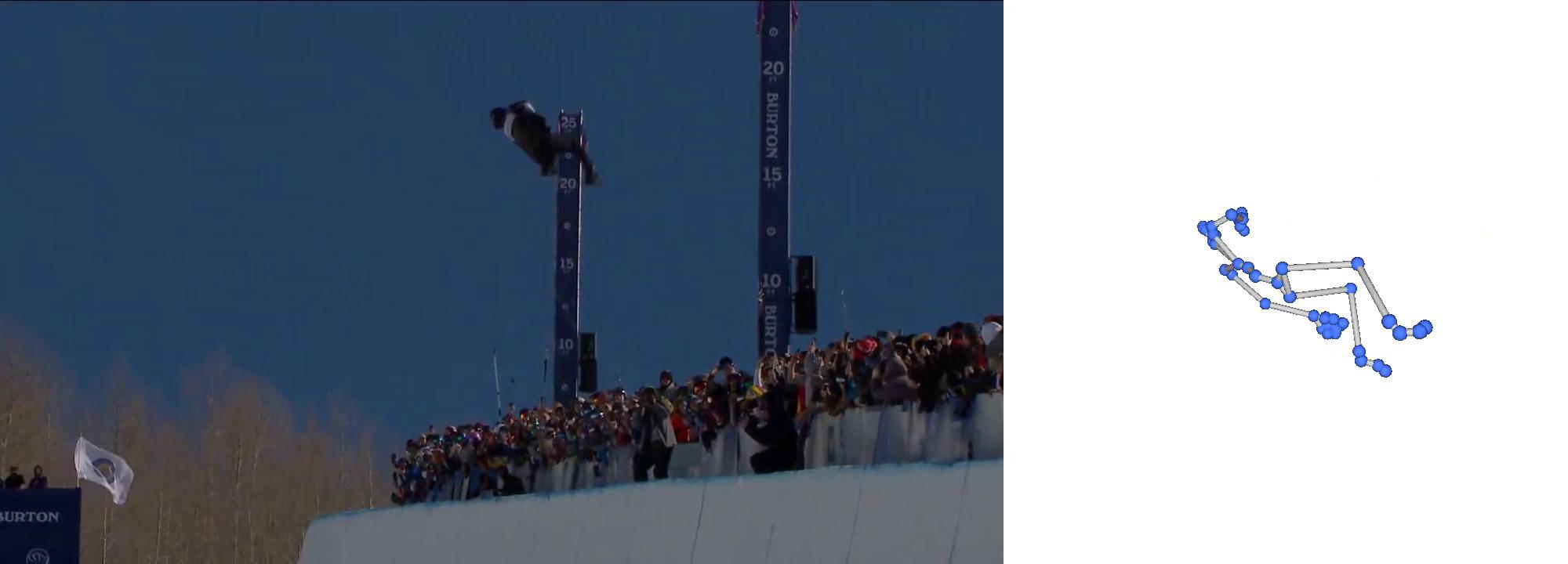} & 
        \includegraphics[width=0.3\textwidth]{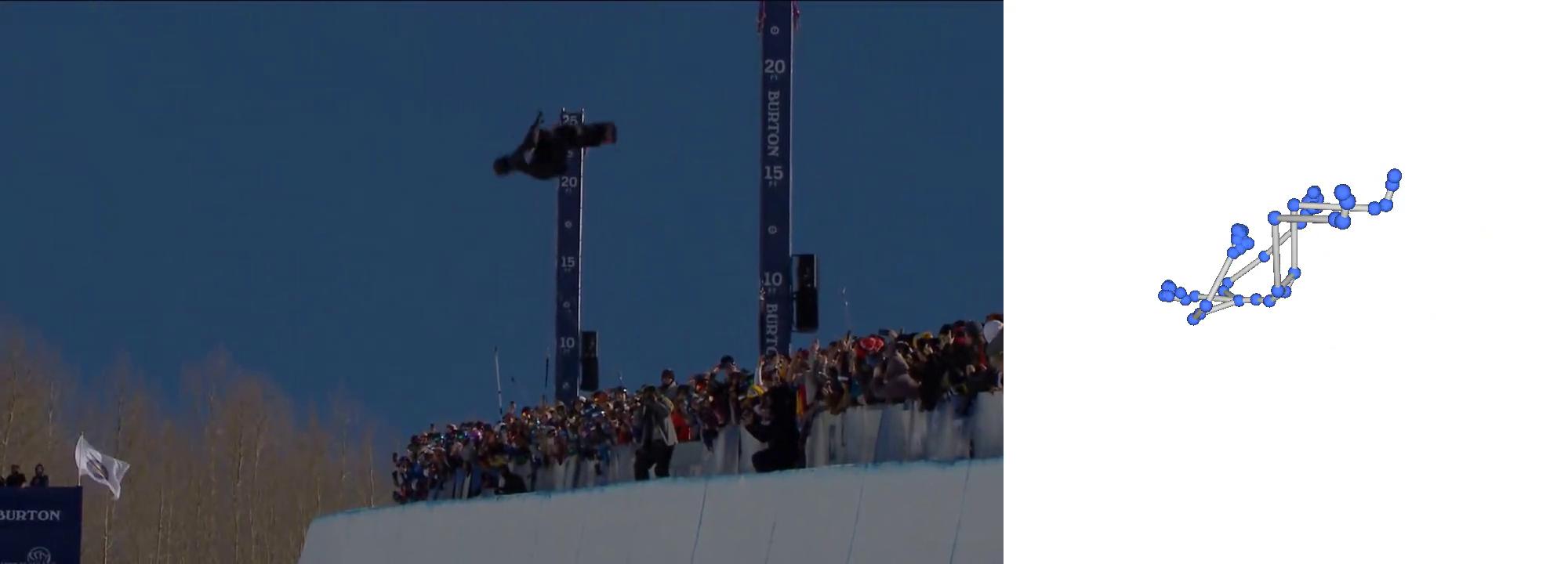} & 
        \includegraphics[width=0.3\textwidth]{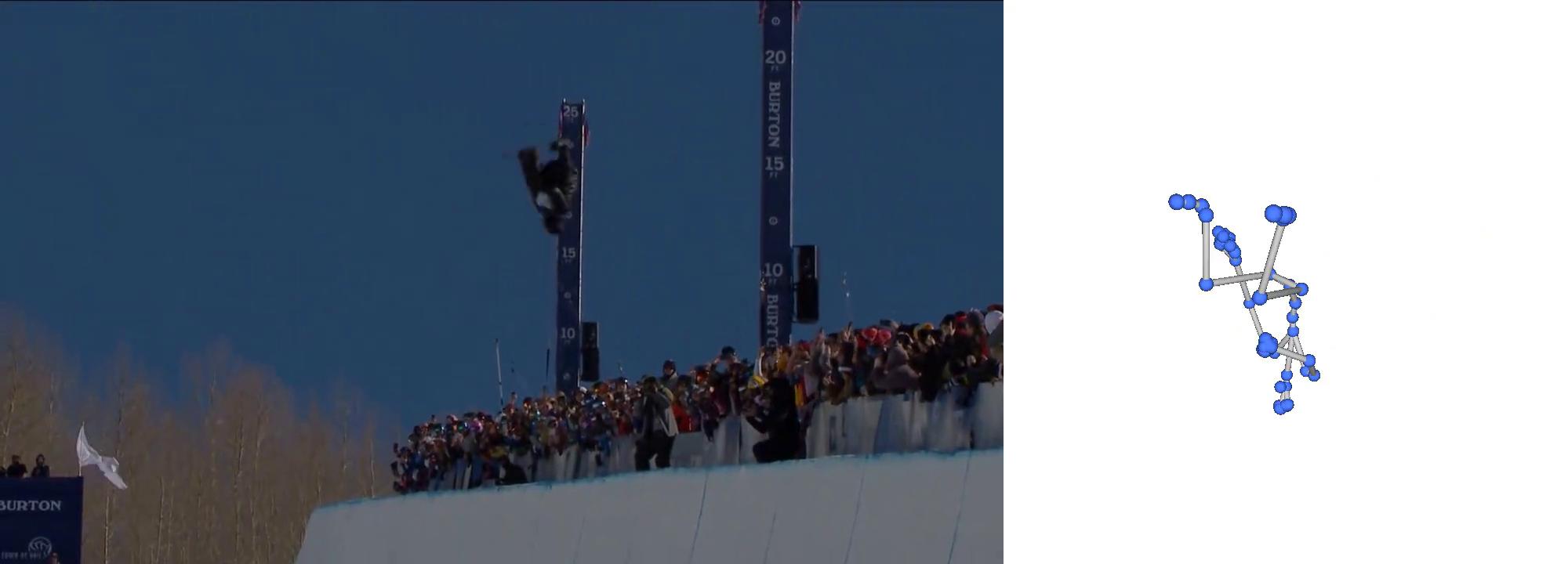} \\
    \end{tabular}
    \captionof{figure}{\textbf{3D Pose Estimation}. Example 3D pose estimation results on challenging skiing (top) and snowboarding (bottom) video. We show every 10-th frame for a subsequence of the input video. Compared to the existing state-of-the-art 3D human pose estimation methods \ModelName takes a full video frame as input and does not require pre-processing steps such as person detection or 2d keypoint estimation.}
    \label{fig:skiing}
\end{table}

\subsection{Implementation Details}
\label{sec: implementation details}

Our approach is implemented based on an open-source and open-weights text-to-video diffusion model WAN 2.1~\cite{wan2025}. 
Following its original architecture, we train our model on video samples at a resolution of 480x832, comprising 81 frames captured at 24 frames per second (FPS). The model's encoder downsamples the input dimensionality with a temporal factor of 4 and a spatial factor of 8 for both width and height.
We train our model with a batch size of 64 on 256 v6e TPUs.
We optimize the model using the Adam optimizer~\cite{kingma2014adam}, with a learning rate of $5e^{-5}$ for 15000 training steps, implementing a linear warmup over the first 250 training steps.
Crucially, for training stability, we apply both gradient clipping, constraining the gradient norm within a certain threshold, and gradient dropping, which discards any batch whose gradient norm exceeds a higher threshold. These techniques proved vital for stable training and high-quality performance.
To enhance training data diversity, we augment our proposed synthetic data with a few publicly available synthetic datasets. Specifically, we incorporate TartanAir~\cite{wang2020tartanair}, Virtual KITTI~\cite{gaidon2016virtual}, and MVS Synth~\cite{huang2018deepmvs} to provide additional depth annotations, while utilizing the camera trajectory data provided by TartanAir~\cite{wang2020tartanair}. For expression-referring segmentation tasks, we leverage real-world datasets due to their immediate availability, integrating MeViS~\cite{ding2023mevis}, Ref-COCO~\cite{yu2016modeling}, and YouTube-VOS~\cite{xu2018youtube}.

\subsection{Inference Cost}
Our feedforward model eliminates the standard WAN's slow 50-step diffusion.
On a single v6e TPU (81 frames, 480x832), the 1.3B model takes 5.92s at 13.6 FPS (DiT 0.96s) and 15.3GB VRAM, while the 14B model takes 10.03s at 8.0 FPS (DiT 5.11s) and 42.8GB VRAM. Both share a 10GB text encoder and 0.25GB VAE, which can be offloaded to system memory to trade speed for VRAM.

\subsection{Evaluation Protocol}
\textbf{{Evaluation Datasets.}} We evaluate our approach on a set of challenging benchmarks that encompass both video and image data. For surface normal estimation, we evaluate on Hi4D datasets~\cite{yin2023hi4d} and SINTEL dataset~\cite{butler2012naturalistic}. For Hi4D, we adhere to the evaluation protocol established in Sapiens~\cite{sapienseccv2024} and DAVID~\cite{saleh2025david}, selecting the same sequences from pairs 28, 32, and 37. These sequences feature six unique subjects captured by camera 4, resulting in a total of 1,195 frames for testing. 
For depth estimation, we evaluate on KITTI~\cite{geiger2013vision}, SINTEL~\cite{butler2012naturalistic}, ETH3D~\cite{schops2017multi}, and Goliath~\cite{martinez2024codec} dataset. For Goliath~\cite{martinez2024codec}, we follow DAVID~\cite{saleh2025david} to perform evaluation to ensure fair comparison, where we utilize the identical selection of 12 cameras, 16 frames, and 4 subjects as in DAVID~\cite{saleh2025david}, which are categorized into data splits of face, upper and full body, with approximately 2.2k frames in total.
For soft foreground segmentation, we report results on VideoMatte~\cite{lin2021real}, PhotoMatte 85~\cite{lin2021real}, and PPM-100~\cite{ke2022modnet}. The original VideoMatte dataset does not provide an official split, thus we evaluate on the static split composite from \cite{lin2022robust}. For camera pose estimation, we evaluate on SINTEL dataset~\cite{butler2012naturalistic}. 
For 3D keypoint estimation, we evaluate on EMDB dataset~\cite{kaufmann2023emdb}.
For expression-referring open-world segmentation, we follow VoCap~\cite{uijlings2025vocap} to reports results on RefVOS-DAVIS~\cite{khoreva2018video} and MeViS~\cite{ding2023mevis}.

\begin{table*}[t]
\centering
\caption{\footnotesize\textbf{Comparison to state-of-the-art specialized models.} We show the performance for a common subset of tasks, including surface normal, depth estimation, camera pose estimation, foreground segmentation, expression-referring open-world segmentation, and 3D human keypoint estimation. "-" denotes that a given model cannot solve the task out-of-the-box, $\sim$ denotes results are not obtained, * denotes results when trained with the same data as our model. Best results are \textbf{bold}, second best \underline{underlined}. Note that on all tasks except for expression-referring segmentation, our method is trained purely on synthetic data. }
\resizebox{\textwidth}{!}{%
\begin{tabular}{@{}ll l cc cc ccc cc cccc c@{}}
\toprule
& \multicolumn{2}{c}{Task}  & \multicolumn{2}{c}{Normals} & \multicolumn{4}{c}{Depth} & \multicolumn{3}{c}{Cam.\ Pose} & \multicolumn{2}{c}{Foreground Seg.} & \multicolumn{2}{c}{Expression-Referring Seg.} & 3D Human \\

\cmidrule(lr){1-3} \cmidrule(lr){4-5}  \cmidrule(lr){6-9} \cmidrule(lr){10-12}  \cmidrule(lr){13-14} \cmidrule(lr){15-16} \cmidrule(lr){17-17}

& \multicolumn{2}{c}{Benchmark} & Sintel~\cite{butler2012naturalistic} & Hi4D~\cite{yin2023hi4d} & Sintel~\cite{butler2012naturalistic} & KITTI~\cite{geiger2013vision} & ETH3D~\cite{schops2017multi} & Goliath~\cite{martinez2024codec} & \multicolumn{3}{c}{Sintel~\cite{butler2012naturalistic}} & V.Mat.~\cite{lin2021real} & P.Mat.~\cite{lin2021real} & Ref-DAVIS~\cite{khoreva2018video} & MEVIS~\cite{ding2023mevis} & EMDB~\cite{kaufmann2023emdb} \\
\midrule

& Method & Backbone & \multicolumn{2}{c}{mAE~$\downarrow$} & \multicolumn{4}{c}{AbsRel~$\downarrow$} & ATE~$\downarrow$ & RPE-T~$\downarrow$ & RPE-R~$\downarrow$ & \multicolumn{2}{c}{MSE~$\downarrow$} & \multicolumn{2}{c}{J\&F~$\uparrow$} & MPJPE~$\downarrow$ \\
\midrule

& David-Large~\cite{saleh2025david} & ViT-L 0.34B  & - & 15.37  & - & - & - & - & - & - & - & - & - & - & - & -  \\
& Sapiens~\cite{sapienseccv2024} & ViT-L 2B & - & {12.14} & - & - & - & - & - & - & - & - & - & - & - & -   \\
& Marigold-E2E-FT~\cite{garcia2025fine} & SD V2 0.86B & 33.5 & $\sim$ & - & - & - & - & - & - & - & - & - & - & - & -   \\
& NormalCrafter~\cite{bin2025normalcrafter} & SVD 1.5B & {30.7} & $\sim$ & - & - & - & - & - & - & - & - & - & - & -  & -  \\
& Lotus-2~\cite{he2025lotus} & FLUX 12B & {30.3} & $\sim$ & - & - & - & - & - & - & - & - & - & - & -  & -  \\

\midrule

& David-Large~\cite{saleh2025david} & ViT-L 0.34B & - & - & - & - & - & {0.012} & - & - & - & - & -  & - & - &  - \\
& Sapiens~\cite{sapienseccv2024} & MAE 2B & - & - & - & - & - & 0.033 & - & - & - & - & - & - & - & -   \\
& MapAnything~\cite{keetha2025mapanything} & DinoV2 1.1B & - & - & 0.306 & 0.096 & $\sim$ & $\sim$ & 0.202 & 0.089 & 2.383 & - & - & - & - & - \\
& VGGT~\cite{wang2025vggt} & DinoV2 1.1B & - & - & {0.247} & {0.067} & $\sim$ & $\sim$ & 0.168 & 0.056 & 0.428 & - & - & - & - & -  \\
& DepthAnything 3~\cite{lin2025depth} & DinoV2 1.15B  & - & - & 0.201 & 0.059 & {0.028} & $\sim$ & {0.065} & 0.048 & 0.456 & - & -  & -  & - & - \\
& D4RT~\cite{zhang2025efficiently} & VideoMAE2 1B & - & - & 0.148 & 0.051 & $\sim$ & $\sim$ & {0.065} & {0.024} & \textbf{0.126} & - & - & - & - & - \\

& VGGT-$\Omega$~\cite{wang2026vggt} & DinoV3 1B  & - & - & ~\underline{0.145} & \textbf{0.041} & \textbf{0.016} & $\sim$ & \underline{0.057} & 0.059 & \underline{0.407} & - & - & - & - & - \\

\midrule
& MODNet~\cite{ke2022modnet} & MNetV2 & -  & - & - & - & - & - & - & - & - & 0.0054 & 0.0304 & - & - & - \\
& RVM~\cite{lin2022robust} & MNetV3  & -   & - & - & - & - & - & - & - & - & \textbf{0.0010} & $\sim$ & - & - & - \\
& David-Large~\cite{saleh2025david} & ViT-L 0.34B & - & - & - & - & - & - &  & - & - & $\sim$ & \underline{0.0009} & -  & - & - \\

\midrule
& ReferFormer~\cite{wu2022language} & Swin-L 0.19B & -  & - & - & - & - & - & - & - & - &  &  & 61.1 & 31.0 & - \\
& VISA~\cite{yan2024visa} & Chat-Univi 13B & -  & - & - & - & - & - & - & - & - & - & - & 70.4 & 44.5 & - \\
& Vocap~\cite{uijlings2025vocap} & SAM2 0.4B & -  & - & - & - & - & - & - & - & - &  & - & 75.1 & 51.9 & - \\
& ReferEverything~\cite{bagchi2025refereverything} & WAN 2.1 14B & - & - & - & - & - & - &  & - & - & - &  & 75.0  & 60.3 & - \\
& SAM 3~\cite{carion2025sam} & PE 0.8B & - & - & - & - & - & - &  & - & - & - & - & 41.3 & 38.3 & - \\
& SAM 3 + Gemini 3.5 Flash~\cite{carion2025sam} & PE 0.8B & - & - & - & - & - & - &  & - & - & - & - & 64.5 & 57.5 & - \\

\midrule

& GVHMR~\cite{shen2024gvhmr} & ViT & - & - & - & - & - & - & - & - & - & - & - & - & - & 72.6 \\
& TRAM~\cite{wang2024tram} & ViT-H & - & - & - & - & - & - & - & - & - & - & - & - & - & 74.4 \\
& Genmo~\cite{genmo2025} & ViT-H & - & - & - & - & - & - & - & - & - & - & - & - & - & \underline{73.0} \\
\midrule
& Diception - Generalist~\cite{zhao2025diception}  & SD3 2B & 37.9 & 20.4 & 0.500 & 0.145 & 0.177 & $\sim$ & - & - & - & - & - & - & - & -  \\

\midrule
& Ours - Specialist - S & WAN 2.1 1.3B & 32.7 & $\sim$ & 0.167 & 0.060 & 0.056 & $\sim$ & \underline{0.057} & \underline{0.023} & 0.720 & $\sim$ & $\sim$ & 69.7 & 59.9 & $\sim$ \\

& Ours - Specialist - L & WAN 2.1 14B & \underline{29.7} & \textbf{10.99} & \textbf{0.130} & \underline{0.048} & {0.037} & \textbf{0.008} & \textbf{0.050} & \textbf{0.018} & 0.464 & \underline{0.0027} & \underline{0.0009} & \textbf{76.4} & \textbf{70.0} & \textbf{71.8} \\

& Ours - Generalist -L & WAN 2.1 14B & \textbf{29.3} & \underline{11.47} & {0.156} & \underline{0.048} & {0.044} & \underline{0.011} & {0.062} & \textbf{0.018} & 0.485 & \textbf{0.0010} & \textbf{0.0008} & \underline{75.8} & \underline{69.0} & $\sim$ \\

\bottomrule
\end{tabular}
}
\label{tab:result_summary}
\end{table*}

\begin{table}[t!]
\centering
\vspace{-0.5em}
\caption{\footnotesize{
Validated on depth estimation, the generative backbone outperform other visual pretraining (V-JEPA and Video MAE v2), and show preliminary data and model scaling properties, where performance improves with data volume and model size
Note that our method was trained exclusively on synthetic data.
}}
\label{tab:depth_comparison}
\resizebox{0.7\columnwidth}{!}{
    \begin{tabular}{@{}l c ccc cc cc cc cc@{}}
    \toprule

\multirow{2}{*}{\textbf{Method}} & \multirow{2}{*}{\begin{tabular}[c]{@{}c@{}}Model\\ Size\end{tabular}} & \multicolumn{3}{c}{Number of Training} & \multicolumn{2}{c}{Sintel} & \multicolumn{2}{c}{KITTI} & \multicolumn{2}{c}{ETH3D} & \multicolumn{2}{c}{Average}  \\

\cmidrule(lr){3-5} \cmidrule(lr){6-7} \cmidrule(lr){8-9} \cmidrule(lr){10-11} \cmidrule(lr){12-13}
 & & Datasets & Videos & Frames & AbsRel$\downarrow$ & $\delta_1$$\uparrow$ & AbsRel$\downarrow$ & $\delta_1$$\uparrow$ & AbsRel$\downarrow$ & $\delta_1$$\uparrow$  & AbsRel$\downarrow$ & $\delta_1$$\uparrow$  \\

    \midrule

    Ours - V-JEPA - H & 0.6B & 1 & 7.5K & 0.9M & 0.422 & 37.3 & 0.226 & 47.5 & 0.196 & 71.8 & 0.281 & 52.2 \\

    Ours - VideoMae V2 - H & 0.6B & 1 & 7.5K & 0.9M  & 0.275 & 37.4 & 0.124 & 64.6 & 0.126 & 84.1 & 0.175 & 62.0 \\

    Ours - VideoMae V2 - G & 1B & 1 & 7.5K & 0.9M & 0.260 & 41.4 & 0.105 & 70.2 & 0.099 & 89.3 & 0.154 & 66.9 \\

    Ours - WAN 2.1 - S & 1.3B & 1 & 7.5K & 0.9M & 0.201 & 72.7 & 0.099 & 89.2  &  {0.068} & {95.6} & 0.122 & 85.8 \\

        Ours - WAN 2.1 - L & 14B & 1 & 7.5K & 0.9M & 0.181 & 76.8 & {0.060} & \underline{97.1} & 0.039 & 98.2 & 0.093 & 90.7 \\
        
    \midrule
    DepthAnything V3 - G~\cite{lin2025depth} & 1.15B & 22+ & 1.23M & $\sim$ 200M & 0.201 & 72.1 & {0.059} & 96.6 & \underline{0.028} & \underline{98.8} & 0.096 & 89.1 \\
    
    D4RT~\cite{zhang2025efficiently} & 1B & 11+ & $\sim$ 1M & $\sim$ 86M & 0.148 & 80.3 & 0.055 & 97.9 & 0.045 & 97.0 & 0.082 & 91.7 \\
    
    VGGT -$\Omega$~\cite{wang2026vggt} & 1B & 33+ & $\sim$ 3M & $\sim$ 600M & \underline{0.145} & \textbf{87.6} & \underline{0.041} & \textbf{98.5} & \textbf{0.016} & \textbf{99.6} & \textbf{0.067} & \textbf{95.2} \\

    Ours - WAN 2.1 - S & 1.3B & 4 & 8.08K & 1.23M & {0.167} & {76.9} & 0.060 & {97.3} & {0.056} & {97.6} & {0.094} & {90.6} \\

    Ours - WAN 2.1 - L & 14B & 4 & 8.08K & 1.23M & \textbf{0.130} & \underline{84.5} & \underline{0.048} & \underline{98.3} & 0.037 & \underline{98.8} & \underline{0.071} & \underline{93.8} \\

    \bottomrule
    \end{tabular}
}
\vspace{-1em}
\end{table}

\noindent\textbf{{Evaluation Metrics.}} 
To assess our model's performance, we report standard metrics appropriate for each task. 
For surface normal estimation, we report the common metrics~\cite{saleh2025david,sapienseccv2024} including mean and median angular error, alongside the percentage of pixels where the error is within the threshold $t \in \{11.25, 22.5, 30\}$. 
For depth estimation, we follow standard practice~\cite{sapienseccv2024,ranftl2020towards,yang2024depth} by adopting the mean absolute value of the relative depth (AbsRel) and the root mean square error (RMSE).
For soft foreground segmentation, we report mean squared error (MSE) as in prior work~\cite{saleh2025david,yang2025matanyone,lin2022robust,lin2021real}.
For expression referring open-world segmentation, we report J\&F scores~\cite{perazzi2016benchmark} (mean of IoU and contour accuracy) averaged on all text queries.
For 3D keypoint estimation, we follow \cite{genmo2025,wang2024tram} to report mean per joint position error (MPJPE).
For camera pose estimation, we report average translation error (ATE), relative pose error - translation (RPE-T), and relative pose error - rotation (RPE-R).

\subsection{Comparison to the state of the art}
We compare our methods with the state-of-the-art models on various challenging benchmarks. Note that our model was trained without using any of the training sets provided alongside the evaluation benchmarks.
Furthermore, our training relies entirely on synthetic data, with the sole exception of the expression-referring segmentation task which incorporates real-world data.
As summarized in Table~\ref{tab:result_summary}, despite its unified and simple architecture, our approach surpasses or delivers highly competitive performance against heavily tailored, task-specific state-of-the-art methods.
For geometric understanding, our method exceeds specialized models like NormalCrafter~\cite{bin2025normalcrafter} and Lotus-2~\cite{he2025lotus} in surface normal estimation, and outperforms or matches dedicated foundation models such as DepthAnything 3~\cite{lin2025depth}, D4RT~\cite{zhang2025efficiently}, and VGGT $\Omega$~\cite{wang2026vggt} in depth estimation and camera pose estimation. In foreground segmentation, our approach achieves competitive results in soft foreground segmentation across both video and image datasets. On expression-referring segmentation, our work show stronger ability in understanding complex sentences compared to SegmentAnything3~\cite{carion2025sam}. Finally, for sparse tasks, it also delivers stronger 3D keypoint prediction performance compared to specialized models including Genmo~\cite{genmo2025} and TRAM~\cite{wang2024tram}.
Qualitative examples are show in Figure \ref{fig:combined_demo}, Figure \ref{fig:segmentation}, and Figure \ref{fig:skiing}.

\subsection{Ablation Studies}

\noindent\textbf{Joint training vs. task-specific training}: 
As demonstrated in Table~\ref{tab:result_summary}, 
transitioning from a specialist to a unified generalist yields mixed behaviors across tasks and benchmarks.
For dense prediction modalities, surface normal estimation shows diver trends on benchmarks, while depth and camera pose estimation generally undergo regression or remain intact (e.g., on KITTI). 
In contrast, foreground segmentation tasks consistently benefit from joint training, whereas expression-referring segmentation remains largely unchanged.

\begin{figure*}[tb]
  \centering
   \includegraphics[width=0.9\textwidth]{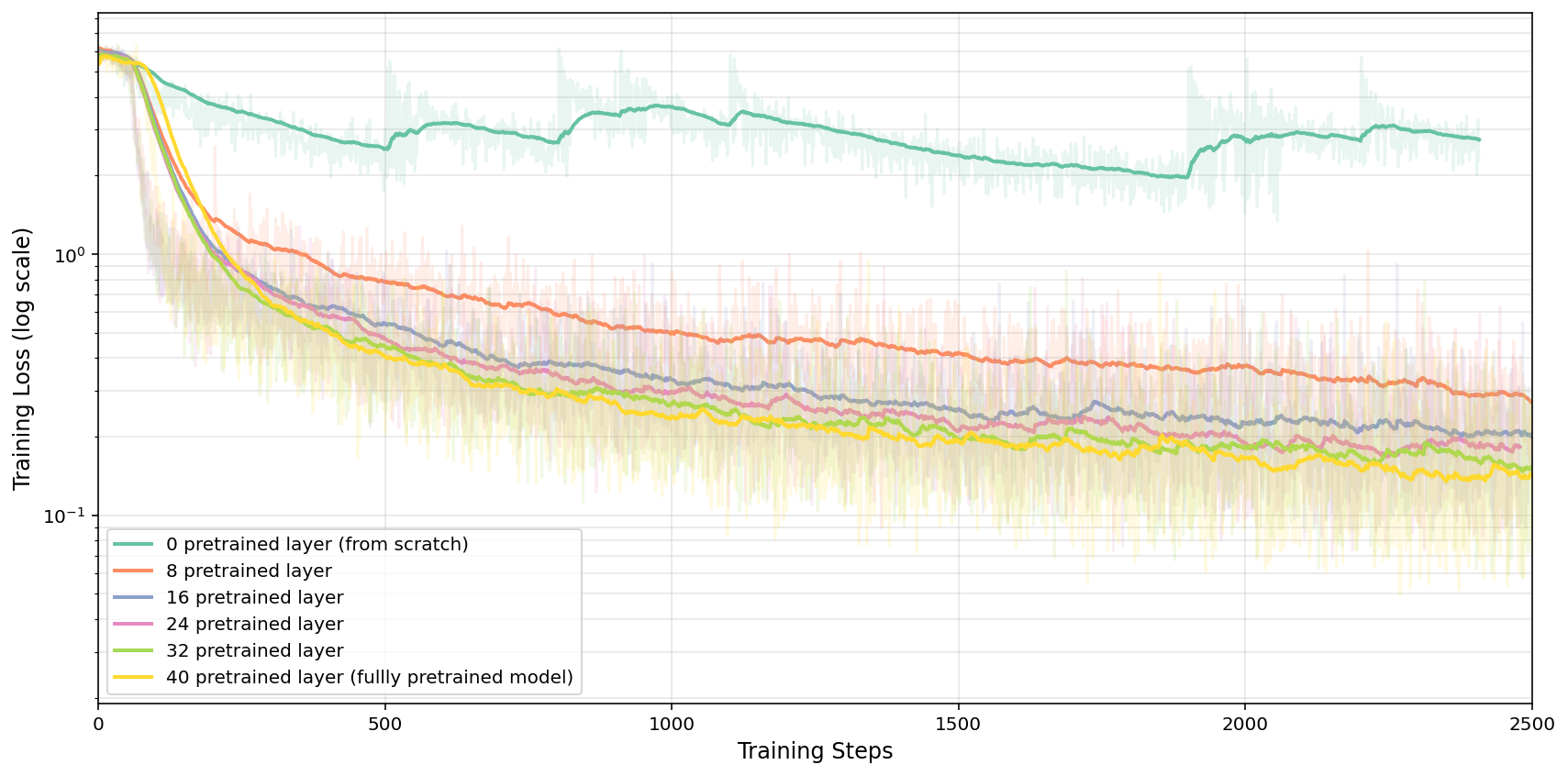}
   \caption{Effect of transferring increasing number of layers from the pre-trained text-to-video model.}
  \label{fig:pretrain transfer}
\end{figure*}
Most notably, we empirically observe that joint training severely degrades 3D human keypoint estimation and compromises other dense tasks. We presume this stems from two architectural factors. 
First, the explicit token-based coordinate regression diverges from the continuous pixel-space pre-training of the generative video model. 
Second, the additional learnable tokens, trained from scratch, disrupt the native attention mechanics of pre-trained DiT layers, damaging their optimized features and requiring more data to converge.
This phenomenon suggest a valuable insight for post-training: one should strive for minimal architectural modifications to the pre-trained backbone, or alternatively, fundamentally redesign the pre-training paradigm to natively accommodate highly diverse downstream tasks.

\noindent\textbf{Importance of pretraining.}
We compare video generative pretraining backbone with other representation learning approaches. 
We use the largest available variants of V-JEPA~\cite{bardes2024revisiting} and VideoMae V2~\cite{tong2022videomae} and fine-tune all on the same training set.
As in Table~\ref{tab:depth_comparison}, our diffusion-based pretraining (WAN 2.1) significantly outperforms other pretraining approaches. 
This serves as a suggestive empirical evidence that the generative diffusion objective itself—rather than just the dataset or model scale—is essential for extracting superior spatiotemporal priors.
Furthermore, VideoMae and V-JEPA lack native support for a unified multi-task model, often requiring task-specific training, while our model can natively use text prompts to steer between tasks.
Furthermore, as in Figure~\ref{fig:pretrain transfer}, training the randomly initialized DiT from scratch yields a nearly flat learning curve. Performance improves as more pre-trained layers are used. This confirms that pre-training is essential for downstream convergence and performance.

\noindent\textbf{Preliminary scaling property and exceptional data efficiency.}
As in Figure~\ref{fig:radar} and Table~\ref{tab:depth_comparison}, the generative backbone demonstrates promising preliminary scaling properties across varying model sizes and datasets, with performance generally improving as model size and data volume increase. The model also shows exceptional data efficiency where it achieves comparable performance with leading models like D4RT and VGGT-$\Omega$ with 7$\times$ to 500$\times$ less training data.

\subsection{Emergent Behaviors}
In addition to competitive performance that matches or surpasses state-of-the-art methods, our model also exhibits intriguing emergent behaviors beyond its intended training objectives. We believe these behaviors arise from the rich 
representations learned in large-scale text-to-video generation models, shedding light on the broader potential of this paradigm for future exploration.

\begin{figure}[t!]
  \centering
  
  \begin{subfigure}[c]{0.49\columnwidth}
    \centering
    \includegraphics[width=\linewidth, trim={0cm 11.75cm 0cm 0cm}, clip]{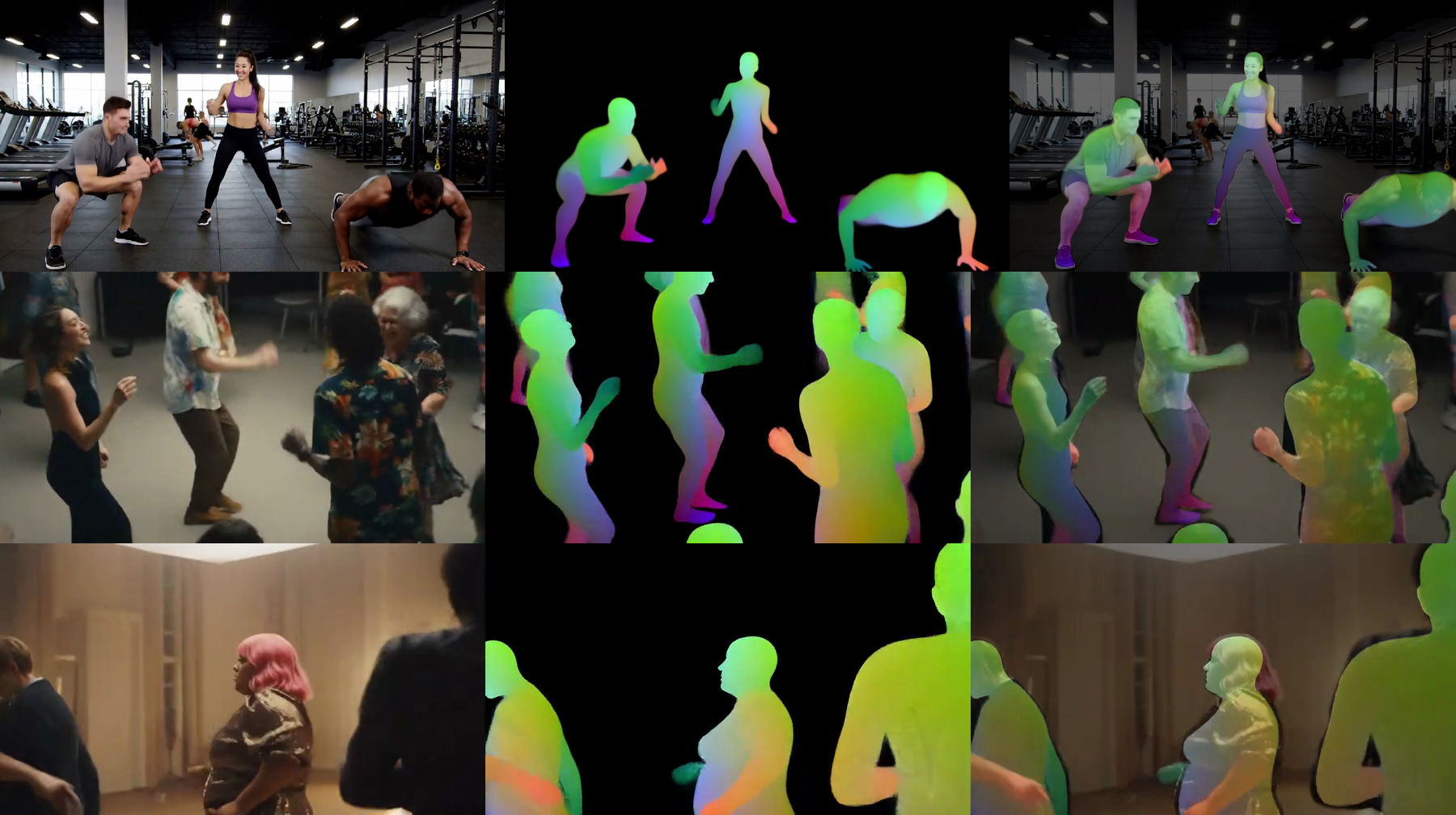}
    \caption{(a) Generalize to multiple instances.}
    \label{fig:multipeople}
  \end{subfigure}
  \hfill 
  \begin{subfigure}[c]{0.49\columnwidth}
    \centering
    \includegraphics[width=\linewidth]{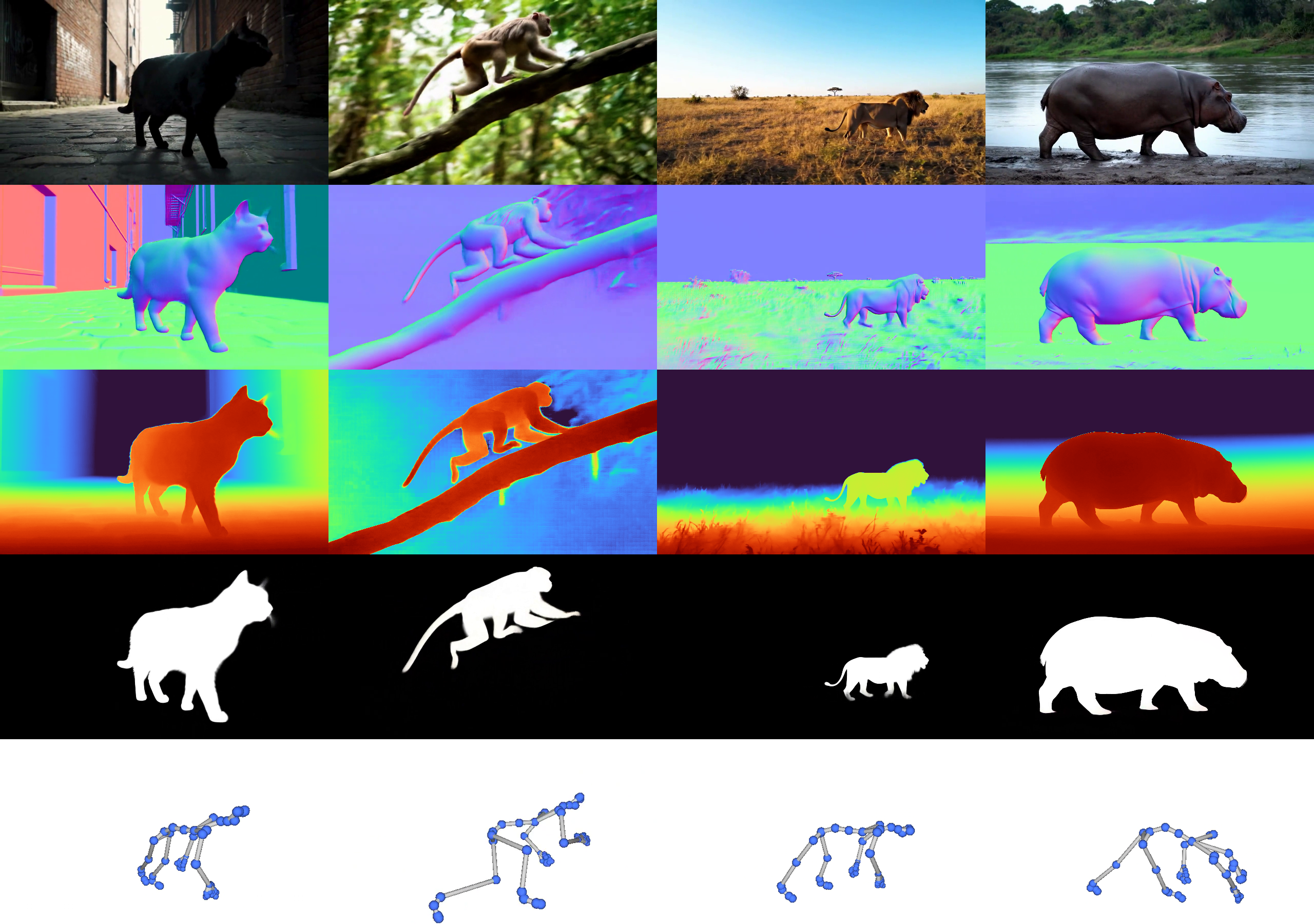}
    \caption{(b) Generalize to unseen categories.}
    \label{fig:oodvis}
  \end{subfigure}

  \caption{\textbf{Generalization capabilities.} Left: Trained on synthetic data with one single object within each video, our method generalizes in zero-shot to real videos with multiple objects. Right: Our approach has been trained only on videos of a single class (humans), but generalizes to a variety of other classes of articulated objects.}
  \label{fig:combined_generalization}
\end{figure}

\noindent\textbf{Generalization from synthetic training data}:
Our approach is trained purely on synthetic videos, yet it generalizes remarkably well to real-world videos.
Quantitatively, as shown in Table~\ref{tab:result_summary} and Table~\ref{tab:depth_comparison}, our model establishes new state-of-the-art performance across all benchmarks. 
Qualatatively, we observe that the fine-grained details present in the output of our model are exceeding the quality of our training data generated with conventional computer graphics methods in Blender, as illustrated in Figure \ref{fig:teaser}, Figure~\ref{fig:combined_demo}, Figure~\ref{fig:segmentation}, and Figure~\ref{fig:skiing}.
For example, see the whiskers on the cat in Figure~\ref{fig:oodvis} (first column) or the soft segmentation of the hair in Figure~\ref{fig:teaser} (second row).

\noindent\textbf{{Generalization to multiple instances}}:
trained on synthetic data with one single object within each video, our method generalizes in zero-shot to real videos with multiple objects as shown in Figure~\ref{fig:multipeople}. 

\noindent\textbf{{Generalization to OOD classes}}:
our approach is trained purely on synthetic videos of human entities, yet it generalizes well to real-world videos—both of humans and of other categories such as animals and anthropomorphic characters—as shown in Figure~\ref{fig:oodvis}.

\section{Conclusion}
We have presented \ModelName, a unified framework that identifies large-scale video generation as a foundational pre-training paradigm for visual perception. By repurposing a pre-trained video diffusion backbone into a high-efficiency, feed-forward model, we demonstrate that the rich spatiotemporal priors and vision-language alignment required for synthesis at large scale can be directly translated into precise perceptual capabilities without the need for costly iterative sampling.
Our experiments across diverse vision tasks confirm that \ModelName not only matches the performance of specialized state-of-the-art models but also exhibits significant emergent behaviors, 
provides strong evidence for the existence of a universal "world model" within generative video backbones. We believe this paradigm shift, from task-specific engineering to scalable generative pre-training, paves a promising path toward a truly unified and generalist vision intelligence for the physical world.
\bibliographystyle{splncs04}
\bibliography{main}


\clearpage

\end{document}